\documentclass[review]{elsarticle}
\usepackage{hyperref}

\usepackage{inconsolata}

\usepackage[namelimits]{amsmath} 
\usepackage{amssymb}            
\usepackage{amsfonts}           
\usepackage{mathrsfs}   
\usepackage{multirow}           
\usepackage{comment}
\usepackage{graphicx} 
\usepackage{bm}
\usepackage{booktabs}
\usepackage{makecell}
  
\usepackage{tabularx}

\usepackage{algorithm}
\usepackage{algpseudocode}

\usepackage{caption}
\usepackage{subcaption}

\usepackage[dvipsnames]{xcolor}
\usepackage{color}

\journal{Elsevier on Oct. 29, 2023}

\begin{document}

\begin{frontmatter}

\title{Low-Resource Court Judgment Summarization for Common Law Systems}

\author[mymainaddress]{Shuaiqi Liu} 
\ead{shuaiqi.liu@connect.polyu.hk}
\author[mymainaddress]{Jiannong Cao}
\ead{csjcao@comp.polyu.edu.hk}
\author[mymainaddress,mysecondaryaddress]{Yicong Li}
\ead{Yicong.Li@student.uts.edu.au}
\author[mymainaddress]{Ruosong Yang}
\ead{rsong.yang@polyu.edu.hk}
\author[mymainaddress]{Zhiyuan Wen}
\ead{preke.wen@connect.polyu.hk}
\address[mymainaddress]{The Hong Kong Polytechnic University, Kowloon, Hong Kong SAR. }
\address[mysecondaryaddress]{University of Technology Sydney, Sydney, Australia.}


\begin{abstract}
Common law courts need to refer to similar precedents' judgments to inform their current decisions. Generating high-quality summaries of court judgment documents can facilitate legal practitioners to efficiently review previous cases and assist the general public in accessing how the courts operate and how the law is applied. Previous court judgment summarization research focuses on civil law or a particular jurisdiction's judgments. However, judges can refer to the judgments from all common law jurisdictions. Current summarization datasets are insufficient to satisfy the demands of summarizing precedents across multiple jurisdictions, especially when labeled data are scarce for many jurisdictions. To address the lack of datasets, we present CLSum, the first dataset for summarizing multi-jurisdictional common law court judgment documents. Besides, this is the first court judgment summarization work adopting large language models (LLMs) in data augmentation, summary generation, and evaluation. Specifically, we design an LLM-based data augmentation method incorporating legal knowledge. We also propose a legal knowledge enhanced evaluation metric based on LLM to assess the quality of generated judgment summaries. Our experimental results verify that the LLM-based summarization methods can perform well in the few-shot and zero-shot settings. Our LLM-based data augmentation method can mitigate the impact of low data resources. Furthermore, we carry out comprehensive comparative experiments to find essential model components and settings that are capable of enhancing summarization performance. 
\end{abstract}


\end{frontmatter}

\section{Introduction}


Common law systems rely on case precedents (prior cases), which encompass not only judgments within a particular jurisdiction but also judgments from all jurisdictions throughout the common law world \cite{commonlawdefdoj}.\footnote{A jurisdiction is an area with a set of laws governed by a court system or government entity \cite{lehmanwest}.} 
Judges in common law jurisdictions need to find similar precedents and refer to the rationale employed in previous judgments \cite{bhattacharya2019comparative}.  Court judgment documents typically contain long text that comprehensively discusses each case and provides detailed explanations of judges' decisions. Reading previous cases' judgments is crucial for legal practitioners in common law jurisdictions. 
There exist massive reported cases, and the number of cases is still increasing \cite{commonlawdefdoj}. It is difficult for legal practitioners to read through abundant cases' judgment documents.

The high-quality summary of judgment document can facilitate readers to quickly browse key information of each case. It not only helps legal practitioners effectively review past cases, but also makes it easier for the general public to read judgments and understand how the courts operate and how the law is applied. 
However, employing legal experts to write summaries costs a lot and has limited coverage of new or atypical cases' judgments \cite{kanapala2019text}.  
Alternatively, we can leverage automatic text summarization technology to generate summaries for court judgments. 
Considering judges need to compare similar precedents across all common law jurisdictions \cite{commonlawdefdoj}, summaries of court judgments from multiple common law jurisdictions are helpful for comparing and analyzing similar precedents efficiently.

Our research focuses on empowering computers to produce high-quality summaries of multi-jurisdictional judgments even when lacking data and computational resources. 
To accomplish this goal, we must address several challenging issues: 1) the lack of datasets, 2) training supervised summarization models with very limited labeled data, 3) efficiently process long documents and summaries with limited computing resources, 4) accurately and comprehensively assess the quality of generated summaries.

\begin{figure*}
\centering
\includegraphics[width=3.0in]{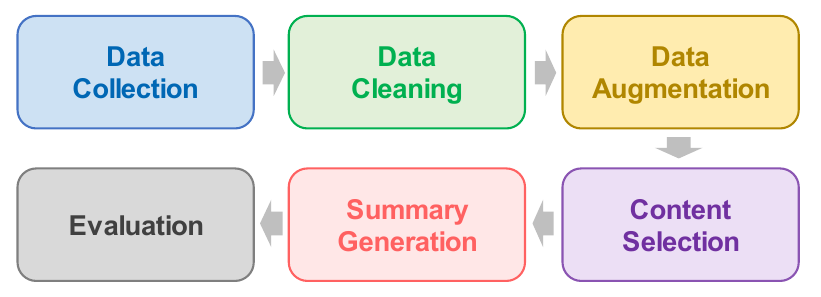}
\caption{Our workflow of Court Judgment Summarization.}\label{fig:clsum_workflow} 
\end{figure*}

We propose a solution for low-resource court judgment summarization to address the above challenging issues. 
Figure \ref{fig:clsum_workflow} depicts six key steps in our solution, including data collection, data cleaning, data augmentation, content selection, summary generation, and evaluation.


There are very few court judgment summarization datasets. Some focus on civil law jurisdictions \cite{de2018rulingbr,glaser2021summarization}, while others concentrate on judgments from specific common law jurisdictions \cite{bajaj2021long,shukla2022legal}. 
They are insufficient to satisfy the demands of summarizing court judgment documents across multiple common law jurisdictions and comprehensively evaluating judgment summarization methods. 
In order to address the lack of datasets, we build CLSum, the first large-scale dataset for summarizing multi-jurisdictional common law court judgment documents. \footnote{The CLSum dataset is available for download online at: github.com/StevenLau6/CLSum.} 
CLSum has four subsets for court judgments from Canada, Australia, the United Kingdom, and Hong Kong SAR.

Similar to other domain-specific tasks, court judgment summarization usually suffers from a shortage of labeled data. Most judgments do not have summaries due to the high cost of hiring experts to write summaries. In most jurisdictions, court websites do not publish or only publish a small number of judgment summaries of typical cases.  
Under this low data resource condition, summarization models' few-shot and zero-shot performance becomes crucial. 
We carry out extensive comparative experiments to analyze the effect of the training set size on different summarization models' performance. 
In addition to selecting models with good few-shot and zero-shot performance, expanding the dataset is another way to improve summarization performance.  
To expand our CLSum's training sets and reduce overfitting, we adopt the large language model (LLM) for data augmentation. Meanwhile, we introduce legal knowledge into the prompts to constrain the LLM to properly use legal concepts when synthesizing sample text in the data augmentation process.


In addition to insufficient data resources, deploying artificial intelligence models in domain applications is often hindered by limited computing resources. Processing long documents and summaries usually requires more computing resources. 
When computing resources are limited, it is crucial to reduce the complexity of the summarization models and improve the efficiency of model training and inference. 
To reduce the complexity of the summarization models, our solution has a two-stage summarization framework, which first compresses long inputs in the content selection stage and then feeds the compressed inputs into the abstractive summarization models for fine-grained content selection and integration. Besides, we also substitute the original self-attention mechanism with sparse attention mechanisms \cite{beltagy2020longformer,dao2022flashattention} and adopt a divide-and-conquer-based training strategy for models pre-trained on the shorter input sequences. 
To further reduce the consumption of GPU memory and conduct model fine-tuning and inference on off-the-shelf GPUs, we adopt some memory-efficient training techniques \cite{dettmers2022llm,wu2023understanding,dettmers2022optimizers,rajbhandari2020zero,dettmers2023qlora,hu2021lora}.

It is also crucial to accurately and comprehensively evaluate the quality of the generated summaries. 
We adopt various summarization methods as baselines and evaluate them on our CLSum dataset. For performance comparison, we carry out automatic evaluation and human evaluation. 
Apart from the widely used evaluation metrics, we design a legal knowledge enhanced evaluation metric named LTScore. It is based on foundation models fine-tuned on the legal corpus. Legal texts usually contain more legal terms compared to texts in the general domain. These terms should be used accurately in court judgments and their summaries. 
Therefore, LTScore assigns greater weights to legal terms in judgment summaries to better assess the accurate usage of such legal terms.

The rest of this paper is organized as follows. We list our objectives and contribution in Section \ref{sec:objs}. Section \ref{sec:relatedwork} briefly introduces related works. We present our CLSum dataset in Section \ref{sec:dataset}, our summarization method in Section \ref{sec:method}, and our experimental settings in Section \ref{sec:experiments}. Our experimental results are reported and analyzed in Section \ref{sec:results_and_discuss}. Finally, Section \ref{sec:conclusionandfuturework} concludes this paper and discusses our future work.

\section{Objectives and Contribution}
\label{sec:objs}
The primary focus of this research is to generate high-quality summaries of judgment documents in the context of insufficient data and computational resources. 
To achieve this goal, we need to complete four objectives:

\begin{itemize}
\item To build a summarization dataset covering court judgment documents from multiple jurisdictions. 
\item To alleviate the impact of insufficient labeled data and improve the summarization model's performance. 
\item To effectively utilize limited computing resources for processing long judgment documents.
\item To accurately and comprehensively evaluate the quality of generated court judgment summaries. 
\end{itemize}

The contribution of this work is threefold:
\begin{itemize}
\item We present CLSum, the first large-scale dataset for summarizing common law court judgment documents from multiple jurisdictions. 
\item We are the first to employ large language models for data augmentation, summary generation, and evaluation in court judgment summarization. 
\item We design a legal knowledge enhanced evaluation metric named LTScore to evaluate generated legal text. 
\end{itemize}

\section{Related Work}
\label{sec:relatedwork}
Automatic text summarization techniques aim to produce concise summaries that retain the salient information within input documents \cite{liu2022long,liu2023neural}. 
Both the publicly available datasets and advanced summarization methods facilitate the summarization research. 
Summarizing short documents (e.g., news and product reviews) has been widely studied \cite{grusky-etal-2018-newsroom,zhuang2006movie,fabbri2019multi,liu-etal-2021-highlight}.  
There has been increasing attention towards long document summarization in recent years. 
Researchers built some summarization datasets for long documents collected from various domains, including academic literature \cite{ijcai2022p591}, government reports \cite{huang-etal-2021-efficient}, and financial reports \cite{liu2022long,liu2023neural}. 

\subsection{Court Judgment Summarization Datasets}
\label{subsec:relatedwork_sum_datasets}
There are very few court judgment summarization datasets. 
Some datasets focus on civil law jurisdictions (e.g., Brazil and Germany) \cite{de2018rulingbr,glaser2021summarization}. While datasets for common law court judgments concentrate on specific jurisdictions. 
The Amicus Briefs dataset \cite{bajaj2021long} focuses on public health cases in the United States\footnote{publichealthlawcenter.org/amicus-briefs}. 
\citet{shukla2022legal} propose datasets summarizing court judgment documents from the United Kingdom and India.\footnote{
India's legal system comprises a blend of common law, civil law, equitable law, as well as customary and religious laws \cite{legalsysindia}.} 

In a common law system, judges are required not only to consider similar local precedents but also to compare similar precedents in other jurisdictions. 
Summaries of court judgments from multiple jurisdictions can help judges compare and analyze similar precedents efficiently. 
However, the formats and content of court judgment documents vary across different jurisdictions. A summarization model that performs well on judgment documents from one jurisdiction may not exhibit effectiveness when applied to judgments from other jurisdictions. 
Current summarization datasets are insufficient to satisfy the demands of summarizing court judgment documents across multiple common law jurisdictions and comprehensively evaluating judgment summarization methods.

\begin{table*}[t]
\scriptsize
\renewcommand\arraystretch{1.0}
\centering
\setlength{\tabcolsep}{0.8mm}{
\caption{Court judgment summarization methods. ``Ext" and ``Abs" stand for the extractive and abstractive summarization methods.} \label{table:relatedwork_sum_methods_compare}
\begin{tabular}{lccccccc}
\toprule 
& \textbf{Type} & \textbf{Models} & \textbf{Jurisdictions} & \textbf{Legal System} \\
\midrule
LetSum \cite{farzindar2004letsum} &  Ext & Heuristic & CA & Common Law \\
CaseSummarizer \cite{polsley2016casesummarizer} & Ext & Heuristic &  Aus & Common Law \\
\citet{saravanan2006improving} & Ext & CRF & India & Mix \\
\citet{de2018rulingbr} & Ext & \makecell*[c]{Gensim, LexRank, \\TextRank} & Brazil & Civil Law \\
\citet{glaser2021summarization} & Ext & CNN, RNN & Germany & Civil Law \\
LegalSumm \cite{feijo2023improving} & Abs & Transformer & Brazil & Civil Law \\
\citet{bajaj2021long} & Abs & BART & US & Common Law \\
\citet{shukla2022legal} & Abs & BART, Pegasus, LED & UK, India & Mix \\
\hline
Ours & Abs & LLMs & HK, CA, UK, Aus & Common Law \\
\bottomrule
\end{tabular}}
\end{table*}

\subsection{Court Judgment Summarization Methods}
\label{subsec:relatedwork_sum_methods}
Document summarization methods can be generally categorized into two types: extractive \cite{erkan2004lexrank,mihalcea2004textrank,li2023mrc} and abstractive \cite{zhang2020pegasus,liu2022key} summarization methods. 
We summarize and compare previous court judgment summarization methods in Table \ref{table:relatedwork_sum_methods_compare}. 

Extractive summarization methods are widely utilized in legal document summarization. They select the most salient input sentences to comprise summaries. 
LetSum \cite{farzindar2004letsum} aims to produce structured summaries comprising four pre-defined themes. It uses heuristic rules to score and rank relevant sentences, and then select the sentences with the highest scores for each theme to comprise the structured summary. 
Similarly, \citet{saravanan2006improving} employ the Conditional Random Field (CRF) to divide a legal document into seven labeled components. 
Then, they use the k-mixture model to select sentences for each component in the structured summary. 
CaseSummarizer \cite{polsley2016casesummarizer} ranks input documents' sentences based on their tf-idf scores \cite{salton1988term} coupled with some customized features. 
\citet{de2018rulingbr} compare some commonly used unsupervised extractive summarization methods (Gensim \cite{vrehuuvrek2010software}, LexRank \cite{erkan2004lexrank}, and TextRank \cite{mihalcea2004textrank}). 
\citet{glaser2021summarization} employ convolutional neural networks (CNN) and recurrent neural network (RNN) based extractive summarization methods. 
Despite the notable development of extractive summarization methods over the last few decades, the extracted summaries still face difficulties in terms of coherence and readability \cite{wang2016exploring,yao2017recent}. Consequently, abstractive summarization methods have received more attention in recent years. 

Abstractive summarization methods identify the salient content in input text and generate novel sentences as summaries \cite{liu2022key}. 
Unlike extractive summarization methods, there are relatively few abstractive summarization methods for court judgment documents. 
LegalSumm \cite{feijo2023improving} divides the legal document into chunks and employs a transformer model \cite{vaswani2017attention} to generate these chunks' summaries. 
Then, it uses a BERT model \cite{devlin2019bert} to assess these chunk summaries' faithfulness and select the one with the highest score as the final summary. 
Its output summaries are much shorter than the target summaries in our CLSum dataset. 
\citet{shukla2022legal} employ some pre-trained sequence-to-sequence models to summarize court judgments. 
\citet{bajaj2021long} fine-tune the BART model \cite{lewis2020bart} on public health judgments in the United States.

There is a lack of research on comprehensively assessing summarization models' performance on multi-jurisdictional judgments. 
Most jurisdictions lack labeled data, which makes it difficult to train supervised summarization models. 
Previous legal document summarization work usually neglects abstractive summarization models' few-shot and zero-shot performance. 
In addition to the scarcity of labeled data resources, deploying artificial intelligence models in domain applications is often constrained by limited computing resources. 
There is a research gap in court judgment summarization under the constraints of low data and computing resources. 
Additionally, more research is needed to utilize legal knowledge in the abstractive summarization of court judgments. 

\section{CLSum Dataset}
\label{sec:dataset}

Common law court judgment summarization (CLSum) is a large-scale summarization dataset covering court judgments from four common law jurisdictions: Canada, Australia, the United Kingdom, and Hong Kong SAR. 
This section first presents our procedures for collecting and pre-processing data. 
Subsequently, we describe four subsets in CLSum. 
Furthermore, we carry out statistics on CLSum and perform a comparative analysis with other datasets.

\subsection{Collecting and Pre-processing Data}
\label{subsec:data_collect_preprocess}

Court judgment documents usually comprehensively discuss each case and explain judges' decisions. Electronic files of judgments are usually publicly available online. 
In most jurisdictions, court websites do not publish or only publish a small number of judgment summaries of typical cases. 
We collect four jurisdictions' court judgment documents together with their summaries from court websites.

After collecting thousands of court judgments' HTML or PDF files, we parse these files and extract their content. 
Then, we conduct a series of data pre-processing operations, including eliminating noises, eliminating replicated examples and outliers with excessively short content, and splitting three sets for training, validation, and test.

\begin{table*}[t]
\small
\renewcommand\arraystretch{1.1}
\caption{Summarization datasets' statistical information. ``Samples" is the sample number in the dataset. ``Doc" and ``Sum" stand for the input document and target summary. ``Sents" and ``Words" represent the mean number of sentences and words. ``Dens." and ``Cov." are the density and coverage of extractive fragments.} 
\centering
\begin{tabular}{lccccccc}
\toprule \textbf{Dataset} & \textbf{Samples} & \textbf{\makecell*[c]{Sents\\ (Doc)}}& \textbf{\makecell*[c]{Words\\ (Doc)}} & \textbf{\makecell*[c]{Sents\\ (Sum)}} & \textbf{\makecell*[c]{Words\\ (Sum)}} & \textbf{Dens.} & \textbf{Cov.} \\
\midrule
CNN/DM & 312,085 & 39.8 & 810.6 & 3.7 & 56.2 & 3.8 & 0.9 \\
PubMed & 133,215 & 87.5 & 3049.0 & 6.8 & 202.4 & 5.8 & 0.8 \\
arXiv & 215,913 & 205.7 & 6029.9 & 9.6 & 272.7 & 3.8 & 0.9 \\
\midrule
CLSum-CA & 192 & 1,168 & 38,403 & 38 & 748 & 0.8 & 0.5 \\\
CLSum-HK & 233 & 395 & 11,911 & 46 & 1,169 & 9.7 & 0.9 \\
CLSum-UK & 793 & 458 & 16,143 & 41 & 1,241 & 2.4 & 0.7 \\
CLSum-AUS & 1,019 & 630 & 20,485 & 19 & 592 & 1.4 & 0.6 \\
\bottomrule
\end{tabular}
\label{table:stats_dataset_overall}
\end{table*}

\subsection{Description of the CLSum's Subsets}
We collect multi-jurisdictional common law court judgment documents to build the CLSum dataset. 
CLSum consists of four subsets: CLSum-CA, CLSum-HK, CLSum-UK, and CLSum-AUS. 

\noindent\textbf{CLSum-CA} is collected from the website of the Supreme Court of Canada (SCC)\footnote{www.scc-csc.ca/case-dossier/cb/index-eng.aspx}. We collect the case briefs and corresponding judgment documents from 2018 to 2023. 
CLSum-CA has the fewest samples among these subsets. 

\noindent\textbf{CLSum-HK} is collected from the legal reference system\footnote{legalref.judiciary.hk/lrs/common/contactus/contactus.jsp}. 
It covers typical cases from multilevel courts, including the Coroner's Court, Magistrates' Court, District Court (DC), High Court (HC), and Court of Final Appeal (CFA) in Hong Kong. We collect these cases' judgment documents and their press summaries from 2012 to 2023.

\noindent\textbf{CLSum-UK} is the subset focusing on the United Kingdom Supreme Court's judgment documents\footnote{www.supremecourt.uk/decided-cases/}. 
\citet{shukla2022legal} collected British judgment documents and their press summaries from 2009 to 2021. 

\noindent\textbf{CLSum-AUS} is collected from the High Court of Australia's website\footnote{www.hcourt.gov.au/publications/judgment-summaries/2023-judgment-summaries}. We collect Australian judgment documents and their summaries from 2005 to 2023. CLSum-AUS is the subset with the most samples.

\begin{table}[t]
\small
\caption{The percentage of target summaries' new n-grams.}
\renewcommand\arraystretch{1.1}
\centering
\setlength{\tabcolsep}{0.7mm}{
\begin{tabular}{lcccc}
\toprule 
\textbf{Dataset} & \textbf{unigrams} & \textbf{bigrams} & \textbf{trigrams} & \textbf{4-grams} \\
\midrule
arXiv & 15.04 & 48.21 & 71.66 & 83.26\\
PubMed & 18.38 & 49.97 & 69.21 & 78.42\\
CNN/DM & 19.50 & 56.88 & 74.41 & 82.83\\
\midrule
CLSum-CA & \textbf{21.65} & \textbf{58.00} & \textbf{80.90} & \textbf{90.09}\\ 
CLSum-HK & 13.48 & 38.86 & 57.53 & 69.06 \\
CLSum-UK & 15.00 & 36.25 & 53.71 & 64.29 \\
CLSum-AUS & 11.65 & 37.69 & 58.01 & 70.53 \\
\bottomrule
\end{tabular}}
\label{table:stats_dataset_novel_ngrams}
\end{table}

\begin{figure}[t]
\centering
\includegraphics[width=3.0in]{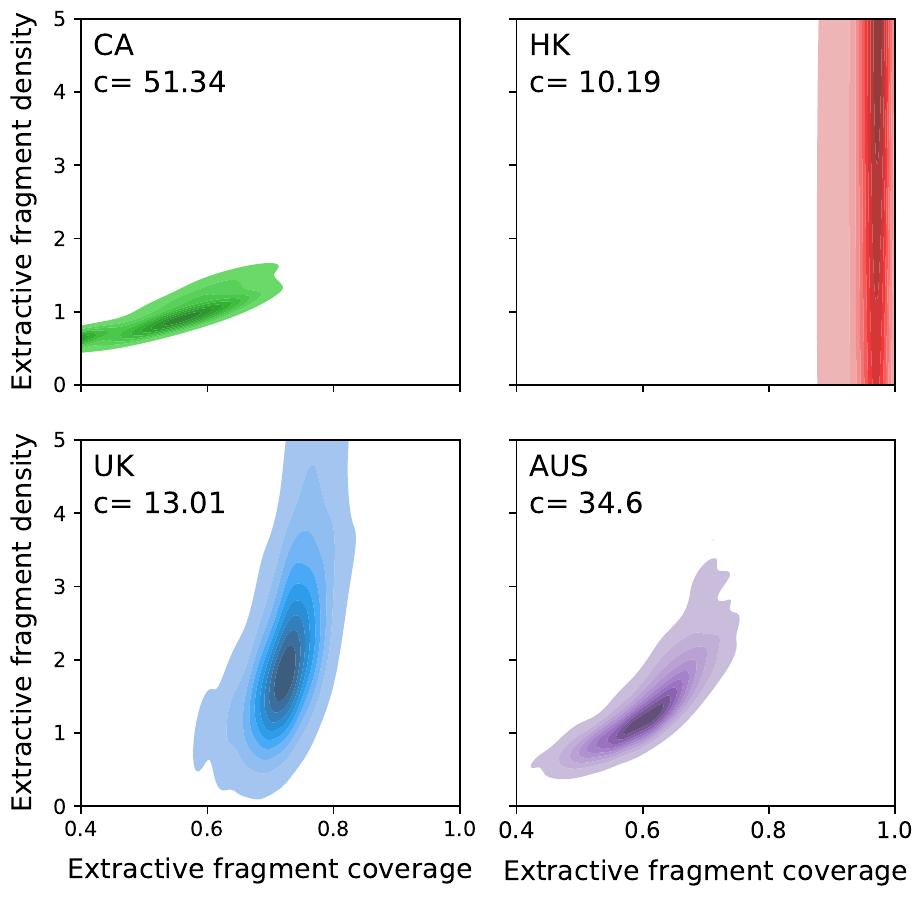}
\caption{Distributions of extractive fragment coverage and extractive fragment density. ``c" denotes the compression ratio.}\label{fig:diversity_visual} 
\end{figure} 

\subsection{Dataset Analysis}
\label{subsec:dataset_analysis}

We conduct statistics on CLSum's four subsets and perform a comparative analysis with other datasets. 
As shown in Table \ref{table:stats_dataset_overall}, these four subsets' input documents and target summaries are much longer in comparison to existing datasets. 
The formats and content of court judgments vary across these four subsets. 
CLSum-HK and CLSum-CA have a few samples, while CLSum-UK and CLSum-AUS have more samples.  
Among these four subsets, CLSum-CA and CLSum-AUS have longer input documents but shorter target summaries.

In order to quantify the abstraction level of CLSum's target summaries, Table \ref{table:stats_dataset_novel_ngrams} counts the ratio of target summaries' n-grams that are not present in the inputs.  
Target summaries of CLSum-CA exhibit a greater number of new n-grams and a higher abstraction level. 
The abstraction level of target summaries in CLSum-HK, CLSum-UK, and CLSum-AUS is comparatively lower than that in other datasets.

Additionally, we utilize two measures \cite{grusky-etal-2018-newsroom}, including the coverage and density of extractive fragments, to assess summarization datasets' extractive nature. 
As shown in Table \ref{table:stats_dataset_overall}, CLSum-HK's coverage is similar to previous summarization datasets but is higher than that of other subsets in CLSum. 
Among these four subsets, CLSum-CA and CLSum-AUS have smaller coverage and density of extractive fragments. 
Figure \ref{fig:diversity_visual} depicts the visualization of distributions of two measures using kernel density estimation. 
CLSum-HK and CLSum-UK subsets have high variability in density, which suggests their target summaries are written in varying styles. 
Furthermore, the compression ratio is calculated by dividing the word count of a document by that of its corresponding summary.

\section{Method}
\label{sec:method}

Figure \ref{fig:clsum_workflow} depicts that our solution consists of six key steps: data collection, data cleaning, data augmentation, content selection, summary generation, and evaluation. 
Our data collection and cleaning procedures are introduced in Section \ref{subsec:data_collect_preprocess}. After the first two steps, we conduct data augmentation to expand the training sets and reduce overfitting to the limited training samples.  
Then, the content selection and summary generation steps complete the selection and integration of key information from rough to fine. 
Our evaluation step comprises both automatic evaluation and human evaluation for assessing the summaries generated by various summarization models.

To the best of our knowledge, this is the first court judgment summarization work adopting LLM in data augmentation, summary generation, and evaluation. 
Summarizing court judgment documents under low data and computing resources conditions has several challenging issues, including: 
training supervised models with extremely limited labeled data, identifying the salient content dispersed within the long judgment document, and improving the efficiency of model training and inference to process long input documents and summaries. 
This section presents our methods to address these aforementioned issues.

\subsection{Mitigating the Impact of Insufficient Labeled Samples}
\label{subsec:method_Insufficient_data}

In most jurisdictions, courts do not publish or only publish a limited number of judgment summaries for typical cases. 
The limited size of the labeled training set usually hinders the performance of supervised models trained from scratch. 
It is essential to guarantee the summarization model's performance when generalizing to cases not seen during training. 
Labeling large-scale datasets can cost a lot, while unlabeled data can be easily collected from the Internet. Researchers pre-trained foundation models with self-supervised tasks on massive unlabeled data to learn better text representations. 
These foundation models can provide good initialization and reduce the amount of labeled training samples required for downstream tasks. 
By fine-tuning on the downstream task, these foundation models often outperform models trained on the same task from scratch. 
To mitigate the impact of insufficient labeled samples from the summarization model perspective, we evaluate diverse foundation models' few-shot and zero-shot performance on our CLSum dataset and select the best-performing model. 

\begin{figure*}
\centering
\includegraphics[width=4.0in]{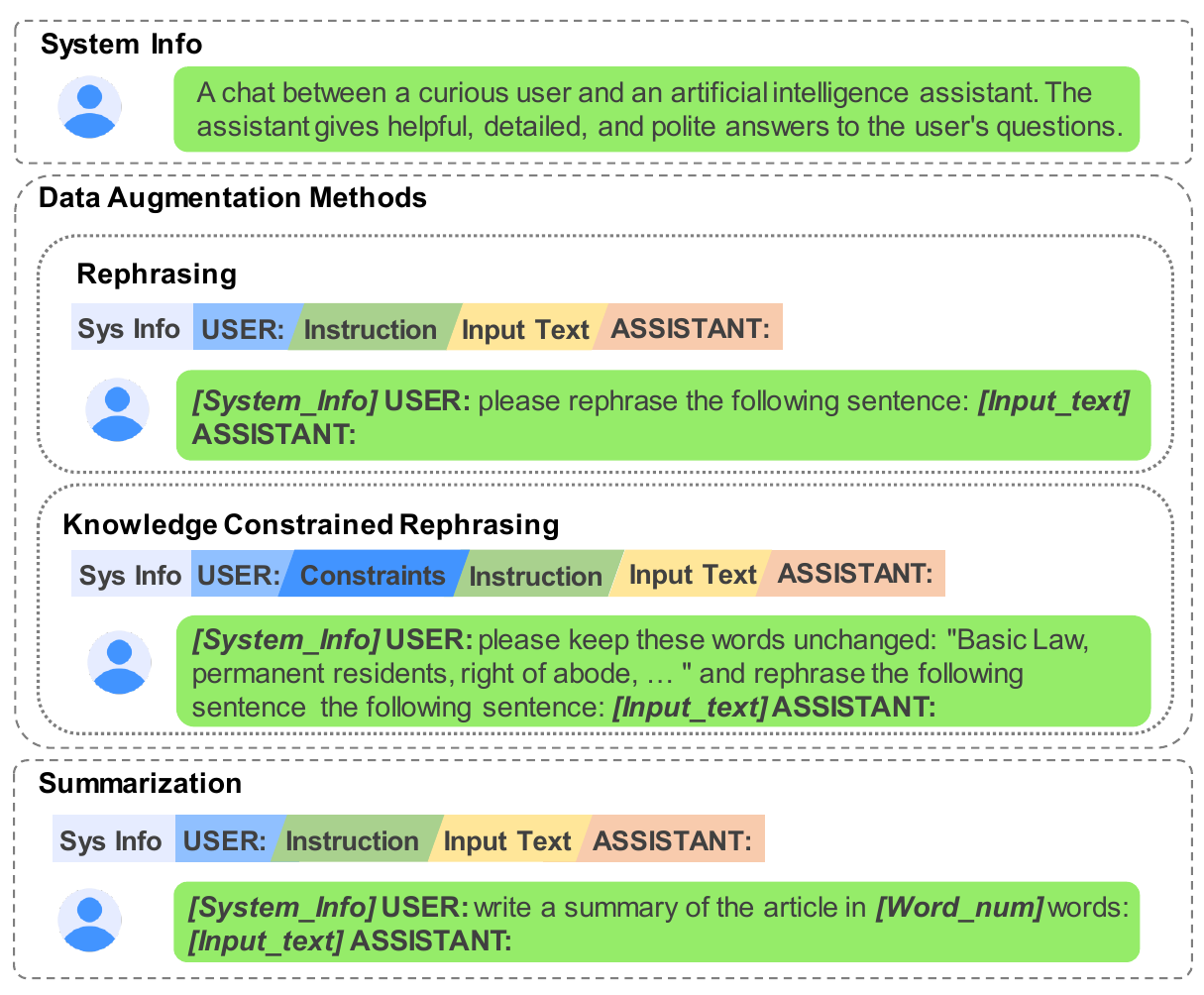} 
\caption{Prompt templates and examples for large language models.}\label{fig:prompt_examples} 
\end{figure*} 

Meanwhile, we also study mitigating the impact of insufficient labeled samples from the data perspective. 
We propose knowledge-constrained rephrasing, an LLM-based data augmentation method constrained by legal knowledge. We also compare it with different data augmentation methods like back translation and rephrasing. These data augmentation methods can expand the training sets and reduce overfitting to the limited training samples.  
The back translation is a commonly used data augmentation method. It first translates the text into another language (e.g., from English to German) and then translates it back to the original language (e.g., from German to English)  \cite{sennrich2016improving}.  
The rephrasing method employs large language models to rephrase each sentence in judgment documents and target summaries \cite{dai2023auggpt}. 
Legal texts usually contain more legal terms compared to texts in the general domain. These terms must be used accurately in court judgments and their summaries. Therefore, we propose knowledge-constrained rephrasing, which introduces legal knowledge into the prompts of LLMs to constrain the synthesized sentences to accurately use legal concepts in the data augmentation process. 
The prompt templates and examples of these data augmentation methods are shown in Figure \ref{fig:prompt_examples}. 
These generated sentences are merged as new data samples. 
We supplement the synthetic data into the training sets to alleviate the impact of insufficient labeled data.

\begin{table*}[t]
\renewcommand\arraystretch{1.1}
\small
\caption{Evaluation results of content selection methods. ``$R_{1}$" is the unigram recall, and ``$R_{avg}$" represents the mean value of the recalls of unigram, bigram, trigram, and 5-gram. ``Lead" represents the truncation method. }
\centering
\setlength{\tabcolsep}{1.0mm}{
\begin{tabular}{lcccccccccccccc}
\toprule
\textbf{\multirow{2}*{Method}} & \multicolumn{2}{c}{\textbf{CLSum-CA}} & & \multicolumn{2}{c}{\textbf{CLSum-HK}} & &\multicolumn{2}{c}{\textbf{CLSum-UK}} & & \multicolumn{2}{c}{\textbf{CLSum-AUS}}\\
\cline{2-3} \cline{5-6} \cline{8-9} \cline{11-12}
~ & $\mathbf{R_{1}}$ & $\mathbf{R_{avg}}$ & & $\mathbf{R_{1}}$ & $\mathbf{R_{avg}}$ & & $\mathbf{R_{1}}$ & $\mathbf{R_{avg}}$ & & $\mathbf{R_{1}}$ & $\mathbf{R_{avg}}$ \\
\hline
Lead & 68.41 & 30.46 & & 85.67 & 52.29 & & \textbf{83.81} & \textbf{54.47} &  & 80.05 & 46.95\\
LexRank & 69.10 & 30.61 & & 85.61 & 52.38 & & 83.15 & 53.84 & & 85.83 & 50.91\\
TextRank & \textbf{69.88} & \textbf{30.88} & & \textbf{85.68} & \textbf{52.45} & & 83.06 & 53.70 & & \textbf{85.90} & \textbf{50.91}\\
\bottomrule
\end{tabular}
}
\label{table:text_segment_selection} 
\end{table*}

\subsection{Salient Content Identification and Integration}
Judgment documents in our CLSum datasets usually contain tens of thousands of words, as depicted in Table \ref{table:stats_dataset_overall}. 
The salient content is dispersed within different parts of these long judgment documents. Nonetheless, foundation models are commonly pre-trained on text sequences that have a predetermined maximum length.  
When abstractive summarization models cannot accept the entire document as input, compressing the input length while preserving key information is crucial. Apart from simply truncating the document, there exist more efficient content selection methods. 

We conduct two-stage operations to identify and integrate the salient content from coarse to fine. 
The first stage, named salient content selection, can be regarded as a rough selection. The recall of essential content that should be retained in summaries is maximized during the compression of long inputs. Subsequently, the condensed inputs are passed to summarization models for fine-grained content selection and integration.

The step of content selection is designed to preserve the maximum key information when compressing the input to a fixed length. 
We compare different methods' performance and mainly focus on their average recall of n-grams. 
LexRank and TextRank methods score and rank input sentences and then select the top-ranked sentences according to a predefined length. \footnote{In our experiments, we employ these content selection methods to compress the input length to 16,384 tokens.} 
Table \ref{table:text_segment_selection} shows these methods' evaluation results on the training set of CLSum. We choose the most effective method for content selection in our subsequent experiments. 
Specifically, we adopt TextRank for CLSum-CA, CLSum-HK, and CLSum-AUS and use truncation for CLSum-UK.

\subsection{Improving the Efficiency of Models and Training Methods}
\label{subsec:efficient_models_training_method}

Most real-world applications not only face low data resources but also have the constraint of low computing resources. 
Especially when available computing resources are limited, improving the efficiency of model training and inference is a crucial issue. 
Many summarization methods require large computing resources when processing long documents, which limits their applications.  
In transformer-based models \cite{vaswani2017attention}, the self-attention mechanism's complexity exhibits a quadratic increase with the input length. 
It can take up a lot of GPU memory and limit models' efficiency. 
Moreover, the limited GPU memory size poses constraints on transformer-based models' capability to model longer context. 
To improve summarization models' efficiency, we employ sparse attention mechanisms \cite{beltagy2020longformer,guo-etal-2022-longt5,dao2022flashattention}. 
Summarization models employing sparse attention mechanisms have the capability to model longer contexts with the same size of GPU memory. 
Additionally, our two-stage summarization framework also reduces the context length that neural summarization models need to model, thus reducing the associated GPU memory consumption. 

In addition to efficient models, efficient training methods can also expedite the training process. 
We adopt some memory-efficient training methods, like gradient accumulation, gradient checkpointing \footnote{github.com/cybertronai/gradient-checkpointing}, parameter quantization \cite{dettmers2022llm,wu2023understanding}, memory-efficient optimizer \cite{dettmers2022optimizers,rajbhandari2020zero}, and adding parameter-efficient adapters \cite{hu2021lora,dettmers2023qlora}. 
For those models pre-trained on the shorter input sequences, we adopt a divide-and-conquer-based training strategy for generating summary segments, followed by merging them to form the final summary. 
These efficient training methods enable us to fine-tune LLMs on lengthy inputs using one off-the-shelf GPU.


\section{Experiments}
\label{sec:experiments}

\subsection{Baselines}
\label{subsec:baselines}
We employ various summarization methods as baselines and evaluate them on our CLSum dataset.  

\noindent\textbf{TextRank and LexRank} \cite{erkan2004lexrank,mihalcea2004textrank}  
are graph-based ranking methods that are widely employed in unsupervised extractive summarization. 

\noindent\textbf{Longformer-Encoder-Decoder (LED)} \cite{beltagy2020longformer} is built on the architecture of the BART model \cite{lewis2020bart} and employs sparse attention mechanisms to replace the original self-attention mechanism within its encoder. 

\noindent\textbf{Legal-LED} \footnote{github.com/nsi319/Legal-Summarizer} is the LED model fine-tuned on the litigation releases of U.S. Securities and Exchange Commission (SEC)\footnote{www.sec.gov/litigation/litreleases.htm}. 

\noindent\textbf{LongT5} \cite{guo-etal-2022-longt5} replaces the self-attention mechanism with a global-local attention mechanism in the encoder part of T5 model \cite{raffel2020exploring} to model longer inputs. 

\noindent\textbf{BLOOM} \cite{workshop2022bloom} is an open-source LLM collection with parameter numbers ranging from 560m to 176B. BLOOM is pre-trained on a corpus comprised of dozens of languages. 

\noindent\textbf{LLaMA} \cite{touvron2023llama} is a collection of LLMs whose parameter numbers range from 7B to 65B. LLaMA is trained on publicly available data. 

\noindent\textbf{Vicuna} \cite{chiang2023vicuna} is a set of LLaMA models fine-tuned with user-shared ChatGPT conversation data.  

\noindent\textbf{GPT-3.5-turbo}\footnote{We adopt the GPT-3.5-turbo-0301 API from Azure Cloud} is the model employed in the ChatGPT. 
Its fine-tuning process employs Reinforcement Learning from Human Feedback (RLHF) on the GPT-3.5 model \cite{IntroChatGPT2022}. 

\subsection{Experimental Setting}
\label{subsec:exp_setting}
For LED-based models (LED and Legal-LED), the vocabulary size is set as 50,265, whereas LLaMA-based models (Vicuna and LLaMA) and LongT5 model utilize a default vocabulary size of 32,000 and 32,128, respectively. 
When fine-tuning the LED-based model, we set the learning rate to $5e^{-5}$. The LLaMA-based models and LongT5 model use $2e^{-5}$ and $1e^{-3}$, respectively. 
We utilize the warmup and decay of the learning rate for all these models. 
As for the optimizer, we use Adam \cite{Kingma2015AdamAM} with $\beta_1=0.9$ and $\beta_2=0.999$ for LED-based models and Adafactor \cite{shazeer2018adafactor} for T5-based models. 
When fine-tuning LLaMA and Vicuna models, we use 4-bit NormalFloat (NF4), QLoRA, and 32-bit paged AdamW optimizer \cite{dettmers2023qlora} to save GPU memory. 
Different foundation models are pre-trained on texts of different lengths. 
In the fine-tuning stage, we predefine the maximum input length for each model to match its input length during the pre-training. 
Given the constraints of GPU memory size, models equipped with sparse attention mechanisms (e.g., LongT5, LED, Legal-LED) can be pre-trained on longer text sequences. Their maximum input length is 16,384. We fine-tuned them to generate the full summary directly. 
For models pre-trained on the shorter input sequences (e.g., LLaMA and Vicuna), we utilize the divide-and-conquer-based training strategy for generating summary segments, followed by combining them to form the final summary. 
We employ the beam search, whose beam size is five.
We utilize the implementations of LongT5, LED, Legal-LED, and 
Vicuna from HuggingFace's Transformers \cite{wolf2020transformers} and LLaMA's implementation from \citet{touvron2023llama}. 
We fine-tune these models using one GPU named Nvidia RTX8000. 

\subsection{Evaluation Metrics}
\label{subsec:evalmetrics}

We carry out automatic evaluation and human evaluation for assessing the summaries generated by various models.  
The automatic evaluation metrics we used can be further divided into statistics-based evaluation metrics (e.g., ROUGE) and model-based evaluation metrics (e.g., BARTScore). We not only employ these commonly used evaluation metrics but also propose a novel evaluation metric named legal text score (LTScore).

We present $\mathrm{F}_1$ scores of Recall-Oriented Understudy for Gisting Evaluation (ROUGE) \cite{lin2004rouge} in our experimental results. 
Specifically, we measure overlaps of unigrams (R-1), bigrams (R-2), and the longest common subsequence (R-L) between output summaries and target summaries.

BARTScore \cite{yuan2021bartscore} is a model-based evaluation metric assessing the quality of generated text by formulating it as a text generation task. 
Built on the pre-trained BART model \cite{lewis2020bart}, BARTScore calculates the log probability of one text sequence y when given another text sequence x. In Eq. \ref{formula:BARTScore}, $\theta$ represents the given pre-trained BART model's parameters. BARTScore sets equal weight $\omega_t$ for each token. 

\begin{equation} 
\operatorname{BARTScore}=\sum_{t=1}^m \omega_t \log p\left(\mathbf{y}_t \mid \mathbf{y}_{<t}, \mathbf{x}, \theta\right) \label{formula:BARTScore}
\end{equation}

Compared with general documents, legal documents usually contain many specialized expressions and domain knowledge. 
Compared with the BART model trained on the general domain corpus, the models trained on legal instruments have a better command of these specialized expressions and domain knowledge. 
To evaluate the generated legal text, we design an evaluation metric named legal text score (LTScore). 
LTScore employs a set of foundation models (e.g., LED and Vicuna) fine-tuned on our legal corpus to vote the score for each text sequence.

Legal texts usually contain more legal terms compared to texts in the general domain. 
These terms must be used accurately in court judgments and their summaries. 
Therefore, LTScore assigns greater weight to legal terms in judgment summaries to better evaluate whether these legal terms are used accurately. 
LTScore can be calculated according to the following formulas.

\begin{subequations} 
\begin{align}
&\mathrm{LTScore}_P=\sum_{j=1}^n \omega_{j}^{'} \sum_{t=1}^m \omega_t \log p\left(\mathrm{cand}_t \mid \mathrm{cand}_{<t}, \mathrm{ref}, \theta_{j}\right) \label{eval_LTScore_p}\\
&\mathrm{LTScore}_R=\sum_{j=1}^n \omega_{j}^{'} \sum_{t=1}^m \omega_t \log p\left(\mathrm{ref}_t \mid \mathrm{ref}_{<t}, \mathrm{cand}, \theta_{j}\right) \label{eval_LTScore_r}\\
&\mathrm{LTScore}_{\mathrm{F}_1}=\frac{2\times \mathrm{LTScore}_P \times \mathrm{LTScore}_R}{\mathrm{LTScore}_P+\mathrm{LTScore}_R} \label{eval_LTScore_f1}\\
&\omega_t =\begin{cases}
1, & \text{if token}_{t} \notin g_{i} \\
1+e^{\omega_{g_i}}, & \text{if token}_{t} \in g_{i} 
\end{cases} \label{eval_LTScore_tokenweight}\\
&\omega_{g_i} = \frac{\mathrm{Score}(g_i)-\mathrm{Score}(G)_{\mathrm{min}}}{\mathrm{Score}(G)_{\mathrm{max}}-\mathrm{Score}(G)_{\mathrm{min}}} \quad \quad g_i \in G \label{eval_LTScore_ngram_weight}
\end{align}
\end{subequations}

Eq. \ref{eval_LTScore_p} and Eq. \ref{eval_LTScore_r} calculate the precision and recall of LTScore. 
In Eq. \ref{eval_LTScore_f1}, the $\mathrm{F}_1$ score of LTScore is the arithmetic average of its recall and precision. 
We adopt a set of foundation models (LED and Vicuna) fine-tuned on our CLSum dataset to vote the final score in Eq. \ref{eval_LTScore_p} and \ref{eval_LTScore_r}. Each fine-tuned foundation model calculates $\log p$ independently. The final LTScore is the weighted sum of the scores calculated by these models. 
$\theta_{j} \in \{\theta_{1},...,\theta_{n}\}$ represents the j-th fine-tuned foundation model's parameters. 
In our experiments, we assign the equal weights $\omega_{j}^{'}=1/n$ for these fine-tuned foundation models.

To emphasize the precise use of legal terms, we assign different weights $\omega_t$ for input tokens. 
For each sample, we select the phrases appearing in the candidate sequence or reference sequence from the glossary of legal terms.\footnote{www.glossary.doj.gov.hk/} 
We rank these selected phrases according to their importance scores and then select the set of phrases $G$ with top-100 importance scores $\mathrm{Score}(G)$. In our experiments, we employ these phrases' tf-idf scores as their importance scores $\mathrm{Score}(G)$. 
The $g_i$ is the i-th phrase in the selected top-100 phrases set $G$. 
Eq. \ref{eval_LTScore_ngram_weight} calculates the Min-Max normalized importance score of $g_i$ as $\omega_{g_i}$. 
Eq. \ref{eval_LTScore_tokenweight} calculates the weight $\omega_t$ of the t-th token $\mathrm{token}_{t}$ in the candidate sequence or reference sequence. If the t-th token $\mathrm{token}_{t}$ is a part of the phrase $g_i$, we add the exponential weight of phrase $g_i$ to the t-th token's weight $w_t$.

LTScore enhances the adaptation to legal texts from two aspects: not only by injecting legal knowledge through fine-tuning base models on the legal dataset but also by adjusting token weights to emphasize the precise use of legal terms. 

\begin{table*}[t]
\renewcommand\arraystretch{0.9}
\scriptsize
\centering
\setlength{\tabcolsep}{0.8mm}{
\caption{\label{autoeval:combined_summary_zero-shot} Automatic evaluation results of summarization models' zero-shot performance.}
\begin{tabular}{lcccc}
\hline
\textbf{\multirow{2}*{Method}} & \textbf{CLSum-CA} & \textbf{CLSum-HK} & \textbf{CLSum-UK} & \textbf{CLSum-AUS} \\
~ & \textbf{R1 / R2 / RL} & \textbf{R1 / R2 / RL} & \textbf{R1 / R2 / RL} & \textbf{R1 / R2 / RL} \\
\hline
LexRank & 31.87/9.54/13.24 & 49.66/23.58/21.41 & 60.28/26.86/22.84 & 53.57/24.46/24.24 \\
TextRank & 32.03/9.36/13.57 & 51.50/24.36/23.65 & \textbf{60.62}/27.22/25.39 & 54.31/24.51/24.61 \\
GPT3.5 & \textbf{50.01}/\textbf{18.58}/\textbf{20.62} & \textbf{54.28}/24.04/\textbf{25.13} & 57.05/25.51/24.10 & 54.10/25.51/25.11 \\
$\mathrm{BLOOM_{560M}}$ & 38.83/11.52/16.41 & 42.31/14.92/19.52 & 49.29/18.15/20.71 & 37.80/14.74/17.84 \\
$\mathrm{BLOOM_{7B1}}$ &  39.05/11.90/16.78 & 46.07/18.10/20.97 & 53.75/22.36/23.15 & 40.29/17.43/19.06 \\
$\mathrm{LLaMA_{7B}}$ & 39.88/11.90/15.99 & 47.60/18.22/20.91 & 54.72/22.52/22.59 & 40.55/15.75/18.88 \\
$\mathrm{LLaMA_{13B}}$ & 40.59/12.63/16.19 & 48.21/18.90/20.88 & 52.80/21.87/22.23 & 43.41/17.44/19.74 \\
$\mathrm{Vicuna_{7B}}$ & 47.32/16.42/20.00 & 53.01/23.20/23.94 & 57.29/27.38/24.40 & 57.27/\textbf{27.53}/\textbf{27.03} \\
$\mathrm{Vicuna_{13B}}$ & 47.69/17.17/20.29 & 53.08/\textbf{24.45}/24.91 & 57.86/\textbf{28.69}/\textbf{25.46} & \textbf{57.33}/27.36/26.75 \\
LongT5 & 23.29/6.08/10.23 & 46.73/16.63/19.32 & 52.32/19.68/20.40 & 43.43/16.93/19.88 \\
$\mathrm{LED_{Base}}$ & 23.63/6.83/11.58 & 47.26/18.16/19.85 & 56.08/21.52/21.43 & 44.84/15.73/21.25 \\
Legal-LED & 37.10/6.97/16.66 & 39.43/9.79/17.88 & 37.46/10.07/17.05 & 42.28/11.08/19.78 \\
$\mathrm{LED_{Large}}$ & 23.87/6.80/10.79 & 47.03/17.27/19.97 & 49.02/17.91/20.85 & 43.64/15.57/20.40 \\
\hline
\end{tabular}}
\end{table*}

\section{Results and Discussion}
\label{sec:results_and_discuss}

In this section, we exhibit our experimental results, and then we analyze and discuss these results. 
We carry out automatic evaluation and human evaluation for assessing the summaries generated by various summarization models. 
Furthermore, we carry out comprehensive comparative experiments to find essential model components and settings that are capable of improving summarization performance. 

\subsection{Summarization Results}
\label{subsec:summarization_results}

We employ multiple metrics to assess the quality of the output summaries in the automatic evaluation. 
Specifically, we adopt the $\mathrm{F}_1$ score of ROUGE \cite{lin2004rouge}\footnote{github.com/bheinzerling/pyrouge}, and some model-based metrics, including $\mathrm{BARTScore}_{F1}$ and our proposed $\mathrm{LTScore}_{F1}$. 
We evaluate different summarization models' zero-shot and few-shot performance.

Tables \ref{autoeval:combined_summary_zero-shot} report the zero-shot performance of different summarization models. 
Under the zero-shot setting, LLMs (GPT-3.5-turbo and Vicuna) are competitive on all subsets of our CLSum dataset. 
The Vicuna models \cite{chiang2023vicuna} fine-tuned with user-shared ChatGPT conversations largely surpass the original LLaMA models \cite{touvron2023llama} and the BLOOM model with similar parameter sizes. 
The zero-shot performance of some pre-trained sequence-to-sequence models with hundreds of millions of parameters (LongT5, LED, Legal-LED, and $\mathrm{BLOOM_{560M}}$) is not as good as that of unsupervised extractive methods (LexRank and TextRank). 

\begin{figure}
\centering
\begin{subfigure}{.49\textwidth}
    \centering
    \includegraphics[width=.99\linewidth]{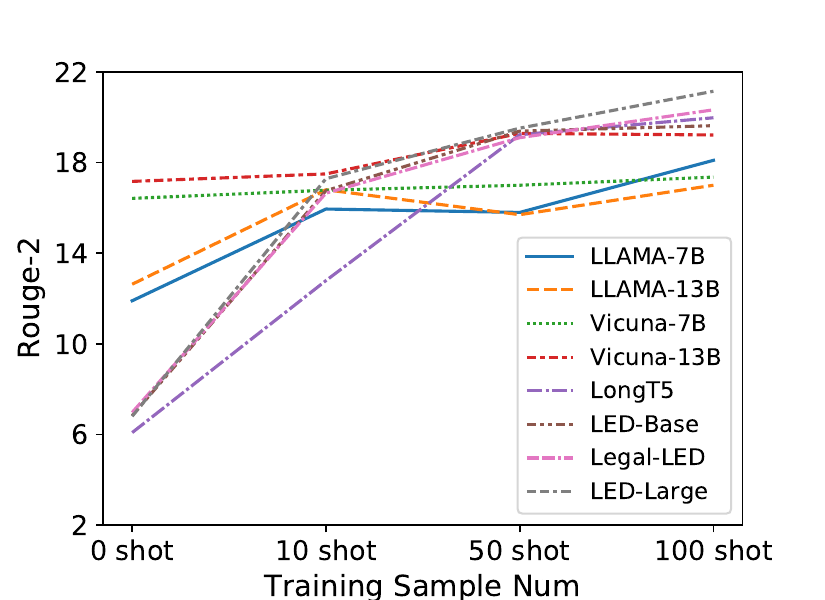}  
    \caption{CLSum-CA}
    \label{fig:clsum_ca_impact_of_training_sample_num}
\end{subfigure}
\begin{subfigure}{.49\textwidth}
    \centering
    \includegraphics[width=.99\linewidth]{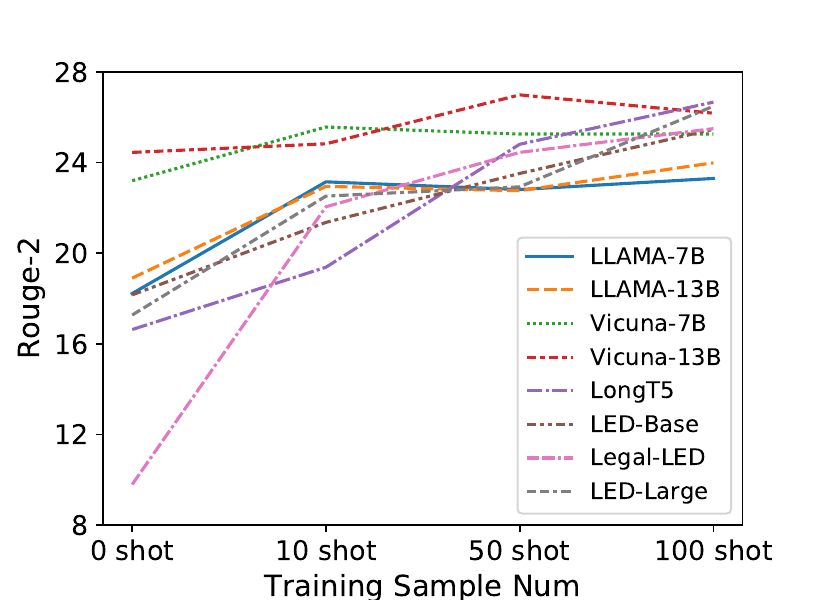}  
    \caption{CLSum-HK}
    \label{fig:clsum_hk_impact_of_training_sample_num}
\end{subfigure}
\begin{subfigure}{.49\textwidth}
    \centering
    \includegraphics[width=.99\linewidth]{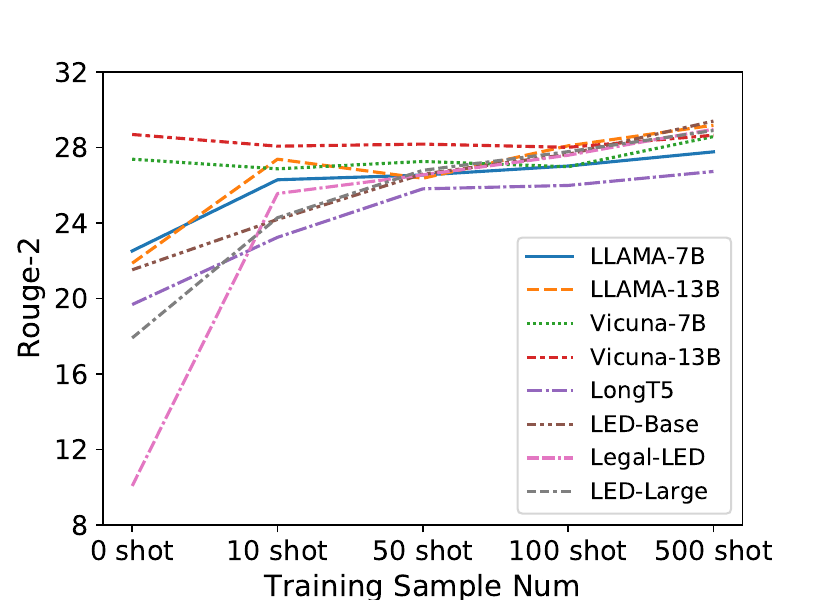}  
    \caption{CLSum-UK}
    \label{fig:clsum_uk_impact_of_training_sample_num}
\end{subfigure}
\begin{subfigure}{.49\textwidth}
    \centering
    \includegraphics[width=.99\linewidth]{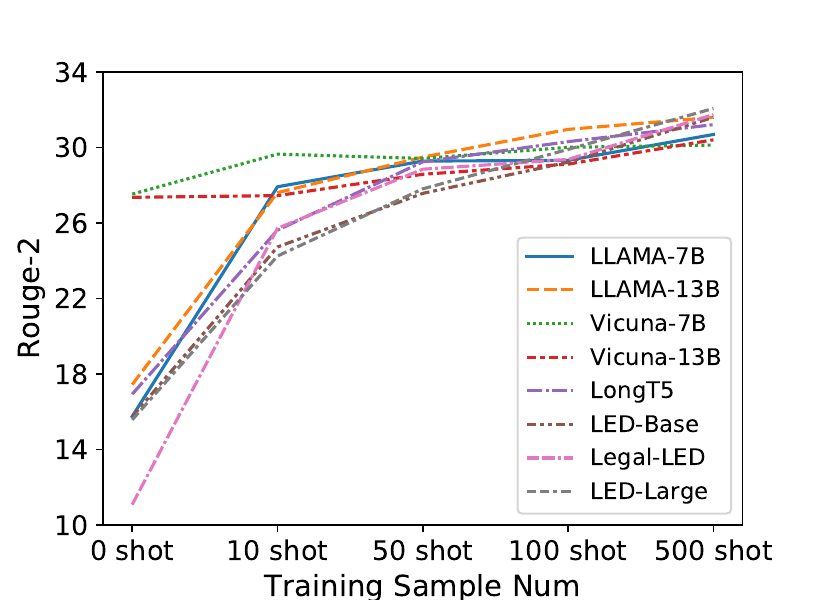}  
    \caption{CLSum-AUS}
    \label{fig:clsum_aus_impact_of_training_sample_num}
\end{subfigure}
\caption{Automatic evaluation result (ROUGE-2 Score) on CLSum.}
\label{fig:clsum_ROUGE2_f1_impact_of_training_sample_num}
\end{figure}

\begin{figure}
\centering
\begin{subfigure}{.49\textwidth}
    \centering
    \includegraphics[width=.99\linewidth]{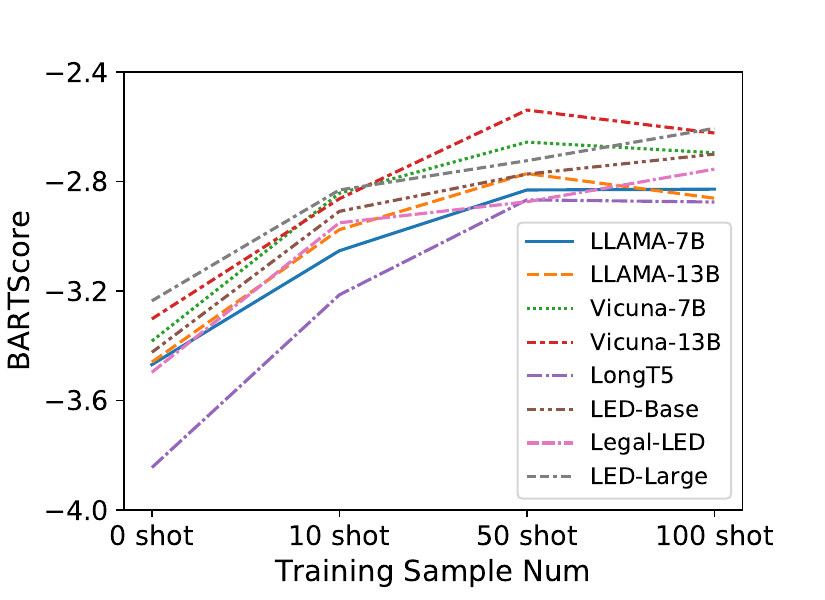}  
    \caption{CLSum-CA}
    \label{fig:clsum_ca_BARTScore_f1_impact_of_training_sample_num}
\end{subfigure}
\begin{subfigure}{.49\textwidth}
    \centering
    \includegraphics[width=.99\linewidth]{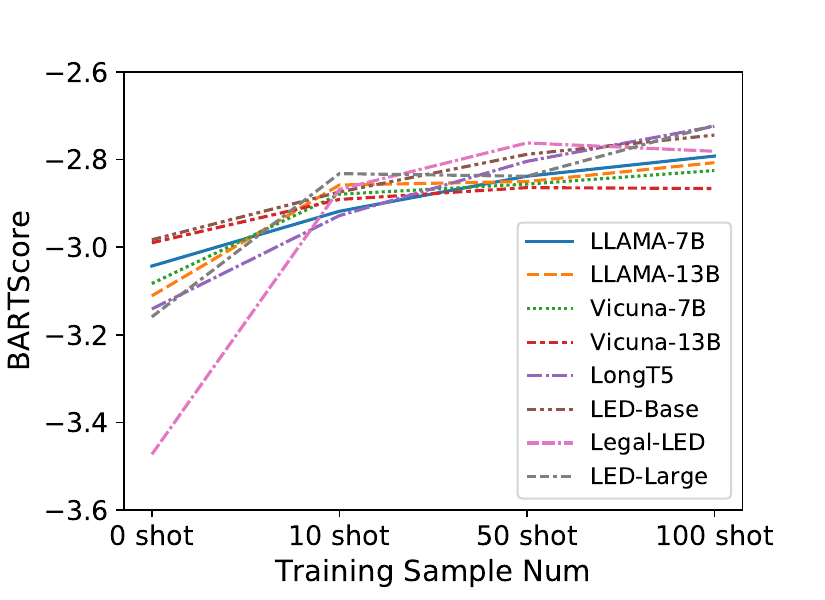}  
    \caption{CLSum-HK}
    \label{fig:clsum_hk_BARTScore_f1_impact_of_training_sample_num}
\end{subfigure}
\begin{subfigure}{.49\textwidth}
    \centering
    \includegraphics[width=.99\linewidth]{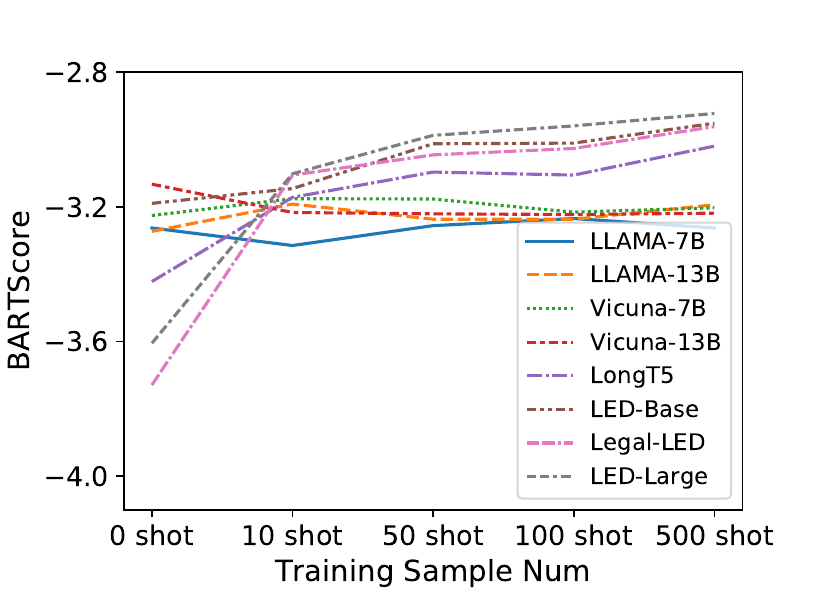}  
    \caption{CLSum-UK}
    \label{fig:clsum_uk_BARTScore_f1_impact_of_training_sample_num}
\end{subfigure}
\begin{subfigure}{.49\textwidth}
    \centering
    \includegraphics[width=.99\linewidth]{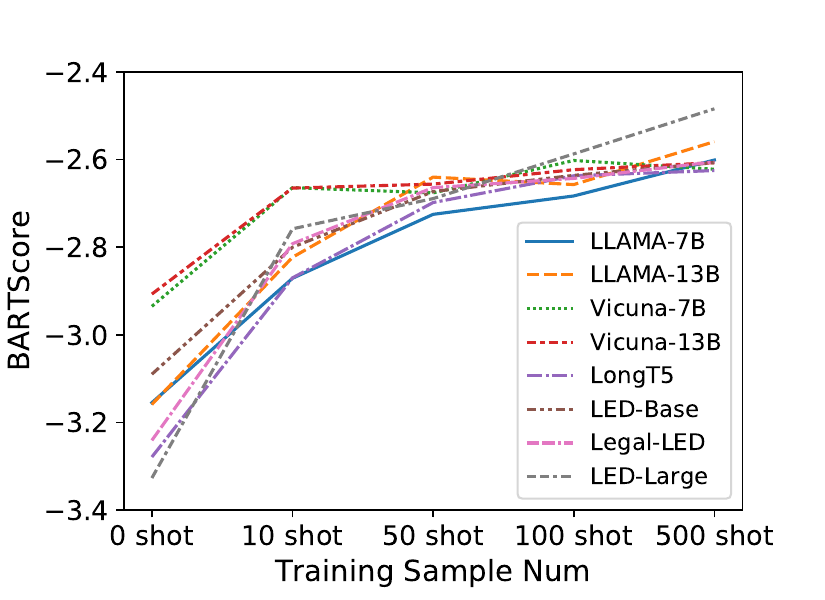}  
    \caption{CLSum-AUS}
    \label{fig:clsum_aus_BARTScore_f1_impact_of_training_sample_num}
\end{subfigure}
\caption{Automatic evaluation result (BARTScore) on CLSum.}
\label{fig:clsum_BARTScore_f1_impact_of_training_sample_num}
\end{figure}

\begin{figure}
\centering
\begin{subfigure}{.49\textwidth}
    \centering
    \includegraphics[width=.99\linewidth]{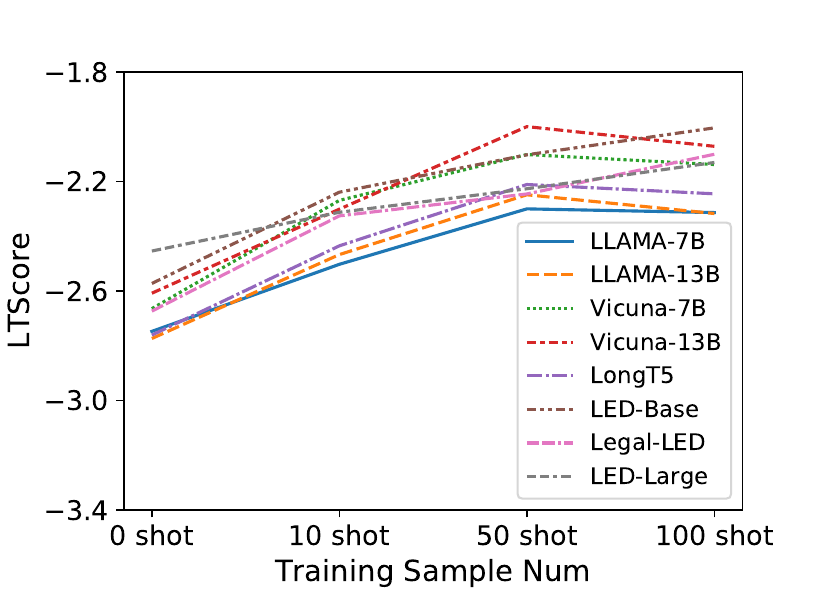}  
    \caption{CLSum-CA}
    \label{fig:clsum_ca_LTScore_average_CA_f1_impact_of_training_sample_num}
\end{subfigure}
\begin{subfigure}{.49\textwidth}
    \centering
    \includegraphics[width=.99\linewidth]{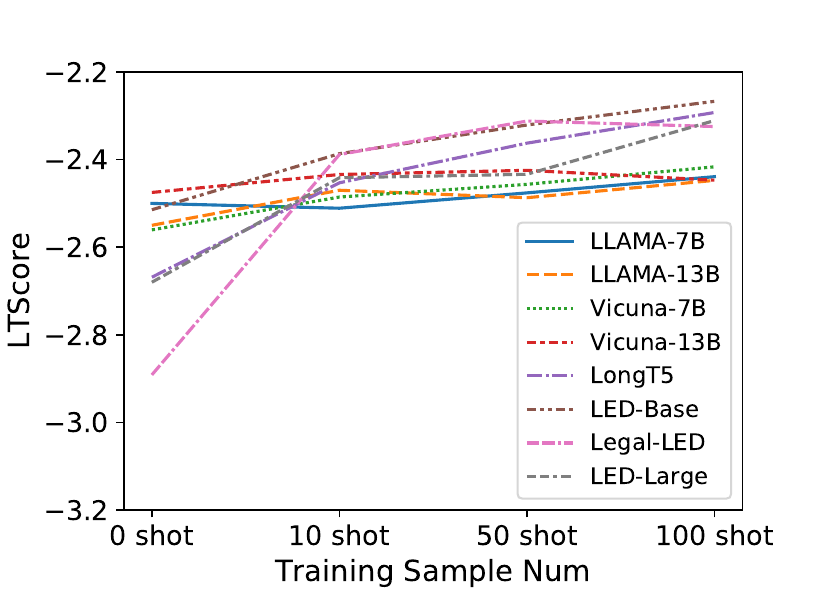}  
    \caption{CLSum-HK}
    \label{fig:clsum_hk_LTScore_average_HK_f1_impact_of_training_sample_num}
\end{subfigure}
\begin{subfigure}{.49\textwidth}
    \centering
    \includegraphics[width=.99\linewidth]{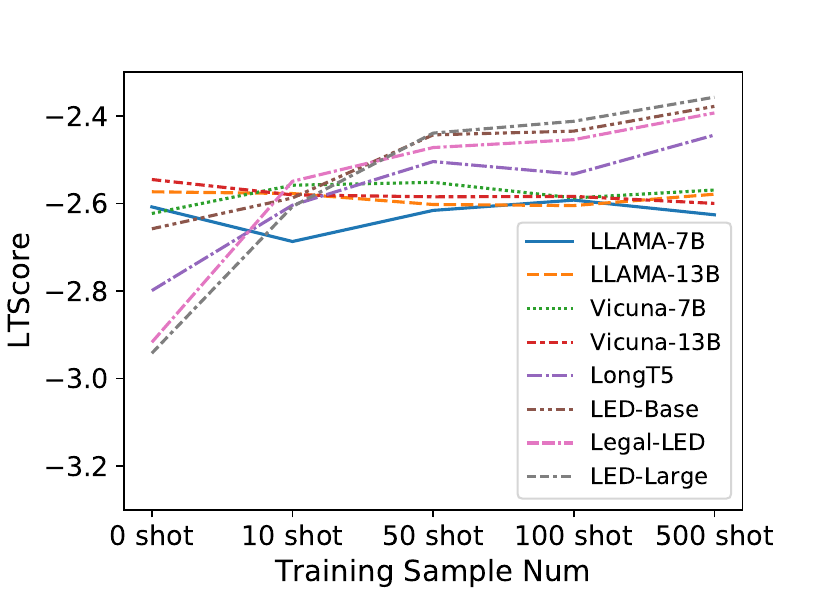}  
    \caption{CLSum-UK}
    \label{fig:clsum_uk_LTScore_average_UK_f1_impact_of_training_sample_num}
\end{subfigure}
\begin{subfigure}{.49\textwidth}
    \centering
    \includegraphics[width=.99\linewidth]{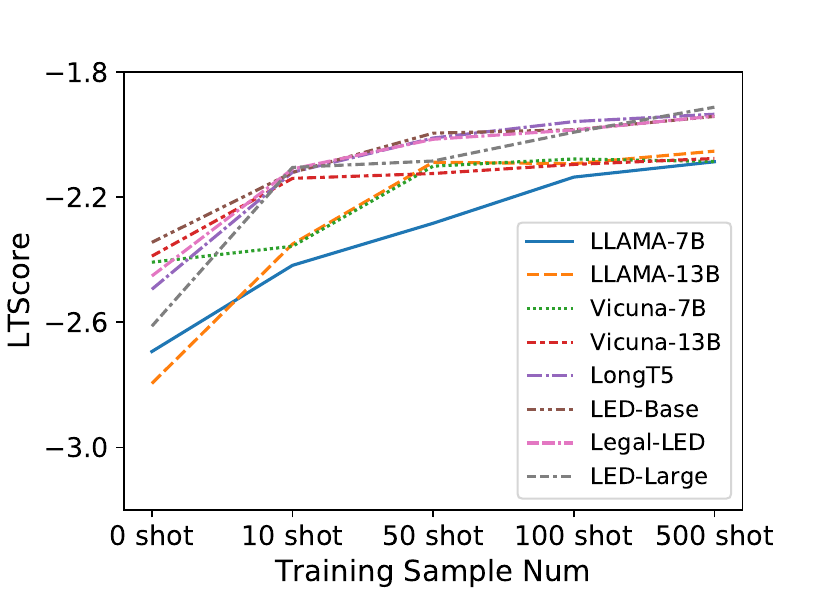}  
    \caption{CLSum-AUS}
    \label{fig:clsum_aus_LTScore_average_AUS_f1_impact_of_training_sample_num}
\end{subfigure}
\caption{Automatic evaluation result (LTScore) on CLSum.}
\label{fig:clsum_LTScoreAverage_f1_impact_of_training_sample_num}
\end{figure}

As for the few-shot setting, we fine-tune summarization models on training sets of increasing size (from ten examples to hundreds of examples). 
Figures \ref{fig:clsum_ROUGE2_f1_impact_of_training_sample_num}, \ref{fig:clsum_BARTScore_f1_impact_of_training_sample_num}, and \ref{fig:clsum_LTScoreAverage_f1_impact_of_training_sample_num} visualize the impact of training set size on the ROUGE-2 scores, BARTScore, and LTScore of the generated summaries.  
Even fine-tuning on only a few examples can bring obvious performance gains for these abstractive summarization methods, which validates the necessity of fine-tuning on downstream tasks. 
These summarization models' performance improvements decay as the training set size increases. 

In our human evaluation, we compare the outputs of summarization models based on their informativeness (i.e., cover salient content of input documents), fluency (i.e., summary content is well organized and uses grammar appropriately), and non-redundancy (i.e., less repetition in output summary). 
We selected 30 samples at random from each CLSum subset's test set. 
For each sample, three annotators assess and compare the anonymously presented output summaries from two models. Additionally, we evaluate the agreement among annotators using Fleiss' kappa \cite{fleiss1971measuring}.

\begin{table*}[t]
\renewcommand\arraystretch{1.1}
\scriptsize
\caption{\label{humaneval:bigbird} Human evaluation results on CLSum dataset. “win” denotes that the current model's output summary surpasses that of $\mathrm{LED_{Large}}$ model in one dimension.}
\centering
\setlength{\tabcolsep}{1.0mm}{
\begin{tabular}{lcccc|cccc}
\hline
 & \multicolumn{4}{c}{\textbf{$\mathrm{Vicuna_{13B}}$}} & \multicolumn{4}{c}{\textbf{$\mathrm{Vicuna_{7B}}$}} \\
\cline{2-5} \cline{6-9} 
~ & \textbf{win} & \textbf{lose} & \textbf{tie} & \textbf{kappa} & \textbf{win} & \textbf{lose} & \textbf{tie} & \textbf{kappa} \\

\hline
\multicolumn{4}{l}{\textbf{CLSum-CA}}& & & &  \\
Informativeness & 23.3\% & 31.1\%& 45.6\%&  0.671 & 21.1\% & 26.7\%& 52.2\%& 0.618 \\
Fluency & 18.9\% & 21.1\%& 60.0\%& 0.623 &  17.8\% & 18.9\%& 63.3\% &  0.603\\
Non-Redundancy & 27.8\%& 36.7\% & 35.6\%& 0.614 & 28.9\%& 33.3\% & 37.8\% & 0.648\\
\hline
\multicolumn{4}{l}{\textbf{CLSum-HK}}& & & &  \\
Informativeness & 24.4\% & 28.9\%& 46.7\% & 0.635 & 26.7\% & 27.8\%& 45.6\% & 0.638 \\
Fluency & 20.0\% & 22.2\%& 57.8\% & 0.634 & 21.1\% & 24.4\%& 54.4\%&  0.629 \\
Non-Redundancy &  17.8\% & 18.9\%& 63.3\%& 0.624 & 18.9\% & 20.0\%& 61.1\% & 0.677\\

\hline
\multicolumn{4}{l}{\textbf{CLSum-UK}}& & & &  \\
Informativeness & 37.8\% & 30.0\%& 32.2\%& 0.648 & 35.6\% & 31.1\%& 33.3\% & 0.649\\
Fluency & 28.9\% & 25.6\%& 45.6\%& 0.655 & 26.7\% & 24.4\%& 48.9\% & 0.612\\
Non-Redundancy & 17.7\% & 20.0\%& 62.2\%& 0.672 & 15.6\%& 16.7\% & 67.8\% & 0.636\\
\hline
\multicolumn{4}{l}{\textbf{CLSum-AUS}}& & & &  \\
Informativeness & 30.0\% & 28.9\%& 41.1\%& 0.662 & 28.9\% & 27.8\%& 43.3\% &  0.625\\
Fluency & 26.7\% & 24.4\%& 48.9\%& 0.647 & 23.3\% & 22.2\%& 54.4\% & 0.629\\
Non-Redundancy & 25.6\% &  26.7\%& 47.8\%& 0.668 & 20.0\% & 22.2\%& 57.8\% & 0.615\\

\hline
\end{tabular}}
\label{humaneval:results}
\end{table*}

\begin{figure}[t]
\centering
\includegraphics[width=4.0in]{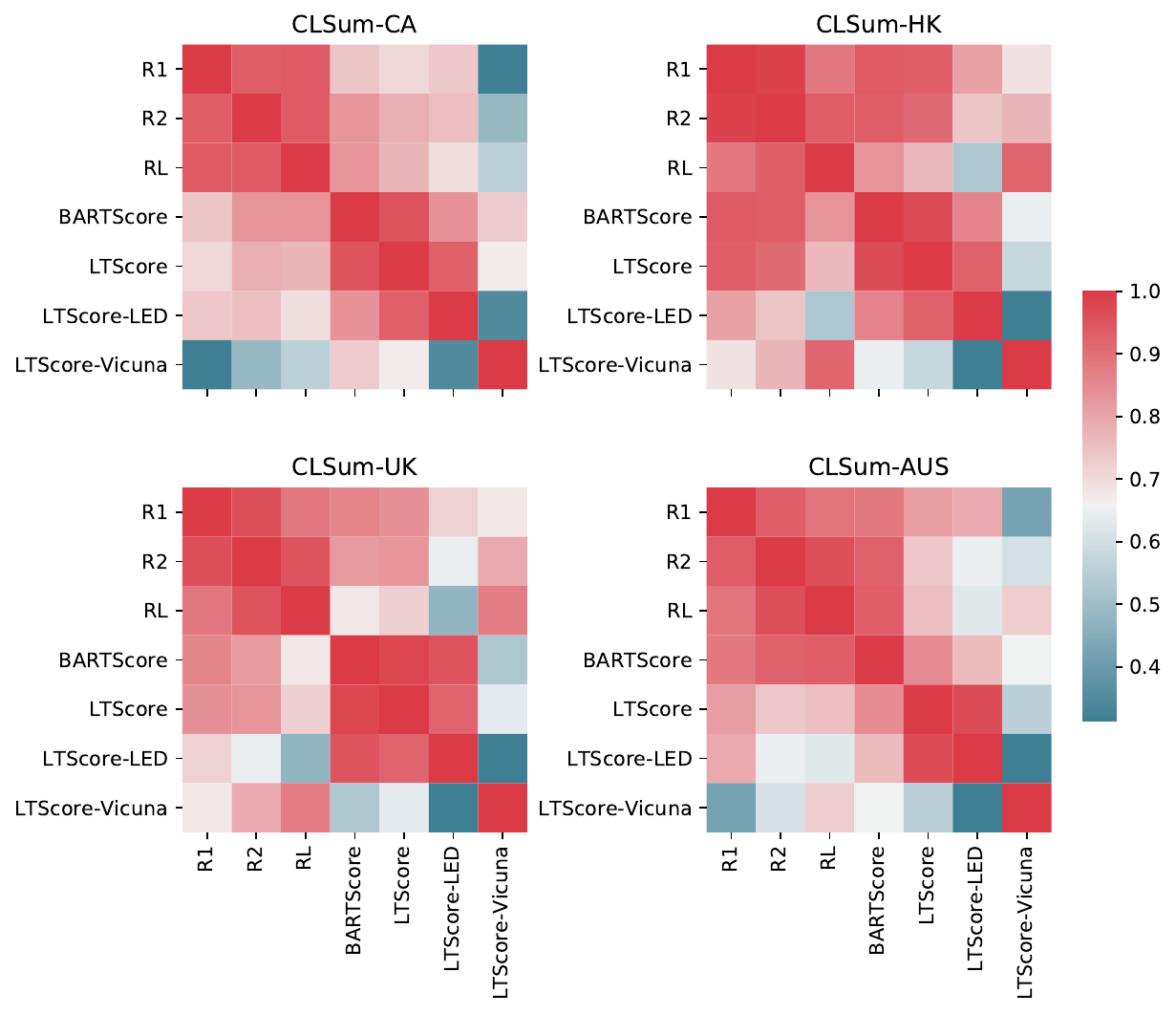}
\caption{Correlation of automatic evaluation metrics. ``LTScore-LED" and ``LTScore-Vicuna" are calculated by a single finetuned foundation model. ``LTScore" denotes the final weighted sum of LTScore. }\label{fig:correl_eval_metrics} 
\end{figure} 

Table \ref{humaneval:results} exhibits our human evaluation results. Three models fine-tuned on each entire subset are compared here. 
The CLSum-CA and CLSum-HK subsets have very few samples in their training sets. On these two subsets, the Vicuna models perform worse than the LED model in terms of informativeness. 
Tables \ref{table:stats_dataset_overall} and \ref{table:stats_dataset_novel_ngrams} present that target summaries' average length is shorter in the CLSum-CA subset. The shorter target summaries in the CLSum-CA comprise more new n-grams that are absent in the input and exhibit lower coverage and density of extractive fragments.  
Generating these more abstractive summaries can be difficult, particularly when the summarization model is fine-tuned on a very small training set. We discover that Vicuna models trained with the divide-and-conquer method on the CLSum-CA subset generate more redundant and less informative summary content than the LED model. 
When there is a lack of training samples, the semantics of the generated summaries for different segments become relatively concentrated and exhibit more repeated content. 
Tables \ref{table:stats_dataset_overall} and \ref{table:stats_dataset_novel_ngrams} also exhibit that the average length of target summaries is longer in the CLSum-HK subset. These longer target summaries' content is more diverse and less abstractive. 
When employing divide-and-conquer, semantically dispersed and longer target summaries for each segment will guide the summarization model to focus on different content when summarizing different segments. 
The lack of training samples in the CLSum-HK subset mainly affects informativeness and has less impact on the redundancy of generated summaries. 
On the CLSum-UK and CLSum-AUS subsets, the Vicuna models can outperform the LED model in informativeness, while these models are comparable regarding fluency and non-redundancy. 
These results verify that the training set size can affect the model acquiring the capacity to effectively summarize key information during fine-tuning, consequently influencing the informativeness of output summaries.

Figure \ref{fig:correl_eval_metrics} depicts the Pearson correlation among various evaluation metrics. 
The correlation among ROUGE scores on different N-grams is high. 
Although the LTScores calculated by the individual fine-tuned foundation model (LED and Vicuna) have a lower correlation with ROUGE scores, the final LTScore (weighted sum) correlated with the ROUGE scores roughly as well as the BARTScore.

\subsection{Discussion on the training set size}

For models trained on the same training set, the larger models' few-shot performance is not always better than that of the smaller models. 
With the increase in the number of training samples, the performance of LLMs (LLaMA and Vicuna) improves slower than these smaller pre-trained sequence-to-sequence models (LongT5, LED, and Legal-LED). 
This may be caused by two reasons: 1) When adopting the QLoRA \cite{dettmers2023qlora} technology, the number of trainable parameters is smaller than the entire model's parameter number. The small number of trainable parameters limits the new knowledge that the model can learn during fine-tuning; 2) Compared with the training data utilized in the pre-training stage and supervised fine-tuning (SFT) stage, the amount of labeled samples used in our fine-tuning process is relatively small, thereby having limited impact on model performance.

Finally, the performance of pre-trained models with hundreds of millions of parameters (LongT5, LED, and Legal-LED) can exceed that of a large language model with billions of parameters (LLaMA and Vicuna). 
Training smaller models with more labeled data can achieve comparable performance, which is critical to reducing the cost of deploying models in real-world applications.

\begin{table*}[!htb]
\scriptsize
\caption{Evaluation results of summarization models trained on augmented datasets.}
\renewcommand\arraystretch{0.8}
\centering
\setlength{\tabcolsep}{0.6mm}{
\begin{tabular}{lcccccc}
\toprule 
\textbf{\multirow{2}*{\makecell{Method}}} & \textbf{\makecell*[c]{Full\\ Train Set}} & \textbf{Rephrasing} & \textbf{\makecell*[c]{Constrained\\ Rephrasing}} & \textbf{\makecell*[c]{Back\\ Translation}}\\
~ & \textbf{R1 / R2 / RL} & \textbf{R1 / R2 / RL} & \textbf{R1 / R2 / RL} & \textbf{R1 / R2 / RL} \\
\midrule
\textbf{CLSum-CA}\\
LLaMA-7B & 47.91/18.10/20.74 & 46.41/16.79/21.24 & 52.17/19.46/22.39 & 39.91/12.06/21.18\\
LLaMA-13B & 48.09/17.00/20.45 & 45.90/17.09/21.56 & 51.02/19.41/22.70 & 47.21/17.30/21.94\\
Vicuna-7B &  47.62/17.36/22.05 & 47.92/17.47/\textbf{23.10} & 52.45/19.65/22.80 & 43.22/14.92/21.48\\
Vicuna-13B &  50.66/19.22/22.68 & 49.93/18.86/23.06  & 51.02/18.49/21.79 & 49.39/18.70/22.36\\
LongT5 & 55.85/19.98/21.48 & 55.01/19.88/21.73 & 55.31/20.18/21.84 & 55.62/19.93/21.70\\
LED-Base & 54.57/19.63/21.32 & 53.28/19.57/21.39 & 54.94/20.08/22.10 & 52.75/19.29/20.80\\
Legal-LED & 56.04/20.33/21.73 & 53.95/19.95/21.63 & 54.95/20.64/22.13 & 55.09/20.29/21.56\\
LED-Large & \textbf{57.23}/21.15/22.65 & 56.64/21.17/22.17 & 56.62/\textbf{21.72}/22.81 & 56.50/21.00/22.28\\
\multicolumn{2}{l}{Average Improvement \%} & -2.14/-1.32/1.65 & \textbf{2.82}/\textbf{4.82}/\textbf{3.29} & -4.62/-6.41/0.20\\
\midrule
\textbf{CLSum-HK}\\
LLaMA-7B & 51.71/23.30/26.18 & 52.14/23.66/25.72 & 53.39/24.04/25.76 & 52.40/23.58/24.83\\
LLaMA-13B & 52.21/23.99/26.06 & 53.15/24.76/26.24 & 53.53/24.67/26.83 & 53.06/23.84/26.06\\
Vicuna-7B &  55.01/25.26/26.42 & 54.32/25.32/26.24 & 54.71/25.27/26.15 &  54.64/24.94/26.07\\
Vicuna-13B &  55.07/26.18/26.78 & 55.30/25.96/26.98 & 56.31/26.45/\textbf{27.02} & 54.87/25.14/26.17\\ 
LongT5 & 56.29/26.67/24.85 & 55.92/26.57/25.14 & 55.80/26.39/25.21 & 55.40/26.10/24.43\\
LED-Base &  55.56/25.47/23.04 & 53.98/24.66/23.36 & 55.89/24.72/24.03 & 55.05/25.51/24.19\\
Legal-LED &  56.10/25.50/23.74 & 55.46/25.82/24.20 & 56.12/25.36/24.85 & 55.88/26.10/24.13\\
LED-Large &  56.43/26.49/24.92 & 56.69/\textbf{27.04}/25.37 & \textbf{57.15}/26.44/25.52 & 55.75/25.85/24.41\\
\multicolumn{2}{l}{Average Improvement \%} & -0.30/\textbf{0.49}/0.66 & \textbf{1.06}/0.29/\textbf{1.76} & -0.27/-0.84/-0.73\\
\midrule
\textbf{CLSum-UK}\\
LLaMA-7B & 60.68/27.65/26.04 & 60.83/28.29/26.68 & 60.47/27.14/25.90 & 60.37/27.37/26.05 \\
LLaMA-13B & 61.13/28.49/26.52 & 60.81/28.21/26.29 & 60.69/28.44/26.75 & 61.27/28.99/27.12 \\
Vicuna-7B &  61.42/29.04/26.83 & 61.58/29.28/27.10 &  61.63/29.46/27.30 & 61.53/28.78/26.77 \\
Vicuna-13B &  61.47/29.15/27.07 & 61.48/29.37/27.38 & 61.27/28.71/27.16 &  61.44/29.24/27.00 \\
LongT5 & 59.62/28.08/26.64 & 59.76/27.81/26.22 & 60.27/28.62/26.86 & 60.54/29.23/26.93\\
LED-Base & 62.18/28.92/25.91 & \textbf{62.63}/29.52/26.58 & 62.41/30.81/27.41 & 61.61/28.49/26.02\\
Legal-LED & 62.59/29.37/25.93 & 62.43/29.51/26.63 & 62.43/30.68/27.57 & 62.05/28.67/25.94\\
LED-Large & 61.55/29.09/26.27 & 61.38/28.80/26.64 & 62.50/\textbf{31.16}/\textbf{28.17} & 61.98/29.41/26.81\\
\multicolumn{2}{l}{Average Improvement \%} & 0.05/0.44/1.11 & \textbf{0.21}/\textbf{2.24}/\textbf{2.82} & 0.04/0.19/0.68\\
\midrule
\textbf{CLSum-AUS}\\
LLaMA-7B & 56.24/28.56/28.71 & 55.87/28.92/29.38 & 57.27/26.07/26.39 & 56.78/28.93/29.29\\
LLaMA-13B & 58.76/31.04/30.64 & 58.69/31.03/30.70 & 57.81/30.02/29.74 & 58.25/31.10/30.73\\
Vicuna-7B &  57.95/30.46/30.77 & 58.18/27.92/27.64 & 57.84/30.10/30.70 & 57.17/29.88/30.33\\
Vicuna-13B &  58.17/30.51/30.86 & 58.10/30.71/30.98 & 57.17/30.30/31.27 & 58.57/31.22/30.96 \\ 
LongT5 & 61.99/31.82/31.55 & 60.90/30.51/30.31 & 61.57/31.52/31.18 & 60.89/30.77/30.74\\
LED-Base & 62.65/32.38/30.92 & 61.75/31.63/30.52 & \textbf{63.14}/\textbf{35.19}/\textbf{33.32} & 62.40/32.14/30.84\\
Legal-LED & 62.42/32.11/30.54 & 61.61/31.70/30.47 & 62.77/34.44/32.61 & 62.27/31.88/30.78\\
LED-Large & 62.64/32.78/31.57 & 62.44/32.38/31.14 & 62.72/32.66/31.27 & 63.07/33.01/31.47\\
\multicolumn{2}{l}{Average Improvement \%} & -0.66/-1.92/-1.76		
 & \textbf{-0.11}/\textbf{0.09}/\textbf{0.31} & -0.29/-0.27/-0.14\\
\bottomrule
\end{tabular}}
\label{table:eval_da_methods}
\end{table*}



\subsection{Discussion on supervised fine-tuning and RLHF}

The Vicuna models \cite{chiang2023vicuna} fine-tuned with user-shared ChatGPT conversations largely surpass the original LLaMA models \cite{touvron2023llama} in the zero-shot setting. 
Regarding the few-shot performance, supervised fine-tuning (SFT) can assist LLMs in achieving commendable results by fine-tuning on a small set of labeled samples. Our experiment results verify the effectiveness of SFT. 

We discover that GPT-3.5-turbo \footnote{We adopt the GPT-3.5-turbo-0301 API from Azure Cloud} fine-tuned with RLHF \cite{IntroChatGPT2022} has difficulty in generating case statements. The RLHF process trains the model to avoid generating illegal content. 
However, court judgment documents usually require an accurate statement of the parties’ illegal facts, which are crucial bases for the judgment. 
RLHF used on general domains is not suitable for legal text generation. It requires a specially designed RLHF process to adequately cater to the complex requirements in the legal field. 
Court judgment summarization models' outputs should accurately and objectively reflect the cases' facts and the court's decisions. There should be no factual or logical errors. Parties from different groups should be treated fairly. 

\subsection{Discussion on data augmentation methods}

The performance of supervised models trained from scratch is typically constrained by the training set size. 
As introduced in Section \ref{subsec:method_Insufficient_data}, we adopt and compare different data augmentation methods, including rephrasing, knowledge-constrained rephrasing, and back translation, to expand the training sets and reduce overfitting to the limited training samples. In our experiments, we doubled the training set size using each data augmentation method. 
Table \ref{table:eval_da_methods} shows the impact of three data augmentation methods on summarization results.  
These data augmentation methods bring different performance gains to summarization models trained on different subsets of CLSum. Experimental results verify that our proposed knowledge-constrained rephrasing method is helpful in the absence of labeled data. 
As shown in Table \ref{table:stats_dataset_overall}, CLSum-CA has the smallest training set. Data augmentation methods bring the most significant performance gain to summarization models trained on this subset. CLSum-AUS has the largest training set. Data augmentation methods bring marginal performance gain to models trained on that subset. 
This verifies that our data augmentation method primarily mitigates the impact of insufficient labeled data. 
The original rephrasing method and back translation method can benefit the ROUGE-L scores, but they often yield negative effects on ROUGE-1 and ROUGE-2 scores. Without the constraints of legal knowledge, there may be many errors in the data synthesized by data augmentation, which can adversely affect the summarization performance. 
Adding constraints in the rephrasing process can ensure the accurate use of legal terms in the synthesized data. This helps train models to accurately use relevant terms when generating judgment summaries.

\begin{table*}[t]
\scriptsize
\caption{Effect of the amount of trainable parameters in the QLoRA adapter.}
\renewcommand\arraystretch{1.1}
\centering
\setlength{\tabcolsep}{1.2mm}{
\begin{tabular}{lccccc}
\toprule 
\textbf{\multirow{2}*{\makecell{Dataset}}} & \textbf{\multirow{2}*{\makecell{Method}}} & \textbf{Rank=8} & \textbf{Rank=16}& \textbf{Rank=32} \\
~ & ~ & \textbf{R1 / R2 / RL} & \textbf{R1 / R2 / RL} & \textbf{R1 / R2 / RL} \\
\midrule
\multirow{4}*{CLSum-CA} & LLaMA-7B & 47.91/18.10/20.74 & 40.35/13.40/20.93 & 42.04/13.09/21.49 \\
~ & LLaMA-13B & 48.09/17.00/20.45 & 48.54/14.42/19.16 &  46.86/16.50/20.91 \\
~ & Vicuna-7B &  47.62/17.36/22.05 & 48.26/17.47/22.49 & 46.22/16.13/22.39 \\
~ & Vicuna-13B &  \textbf{50.66}/\textbf{19.22}/22.68 & 49.66/19.04/22.29 & 50.44/19.14/\textbf{22.84} \\
\midrule
\multirow{4}*{CLSum-HK} & LLaMA-7B & 51.71/23.30/26.18 & 53.73/24.31/26.32 & 53.64/24.25/26.16 \\
~ & LLaMA-13B & 52.21/23.99/26.06 & 54.13/25.31/26.61 &  54.89/25.83/26.90\\
~ & Vicuna-7B &  55.01/25.26/26.42 & 55.02/25.20/26.40 &  55.13/25.82/26.58\\
~ & Vicuna-13B &  55.07/26.18/26.78 & \textbf{55.82}/\textbf{27.25}/\textbf{27.54}&  \textbf{55.82}/26.95/27.00\\
\midrule
\multirow{4}*{CLSum-UK} & LLaMA-7B & 60.68/27.65/26.04 & 61.04/27.88/26.05 & 60.55/27.96/26.21 \\
~ & LLaMA-13B & 61.13/28.49/26.52 & 60.75/28.19/26.34 & 60.52/28.14/26.03 \\
~ & Vicuna-7B &  61.42/29.04/26.83 & 61.45/28.44/26.56 & 61.19/28.80/26.47 \\
~ & Vicuna-13B &  \textbf{61.47}/\textbf{29.15}/\textbf{27.07} & 61.30/28.42/26.71 & 60.93/28.66/26.81 \\
\midrule
\multirow{4}*{CLSum-AUS} & LLaMA-7B & 56.24/28.56/28.71 & 56.21/28.61/28.95 & 56.32/28.55/28.82 \\
~ & LLaMA-13B & 58.76/31.04/30.64 & \textbf{59.15}/\textbf{31.82}/\textbf{31.81} &  58.62/31.35/31.60 \\
~ & Vicuna-7B &  57.95/30.46/30.77 & 56.98/30.14/30.35 & 57.45/30.34/31.00 \\
~ & Vicuna-13B &  58.17/30.51/30.86 & 57.19/30.09/30.94 & 57.88/30.56/30.94 \\
\bottomrule
\end{tabular}}
\label{table:eval_adapter_train_para_num}
\end{table*}

\subsection{Discussion on adapters' trainable parameters}

The adapter is a small set of trainable parameters added to the large language models. 
We use the Low-rank Adapter (LoRA) \cite{dettmers2023qlora,hu2021lora} to reduce the consumption of GPU memory when fine-tuning LLaMA and Vicuna models. 
As shown in Eq. \ref{equa:lora}, LoRA supplements the original linear projection $h=W_0 x$ with an additional factorized projection. During training, $W_0 \in \mathbb{R}^{d \times k}$ remain unchanged, whereas $A \in \mathbb{R}^{r \times k}$ and $B \in \mathbb{R}^{d \times r}$, which have a rank $r \ll \min (d, k)$, comprise trainable parameters.

\begin{equation}
h=W_0 x+\Delta W x=W_0 x+B A x \label{equa:lora}
\end{equation}


Table \ref{table:eval_adapter_train_para_num} shows the impact of LoRA's rank $r$ on summarization results. 
The number of trainable parameters in the adapters expands as the rank $r$ increases.  
Results show that increasing the LoRA's rank $r$ does not necessarily improve the generated summaries. 
The primary constraint on the model performance stems from the inadequate quantity of training samples.

\begin{figure*}
\centering
\includegraphics[width=4.9in]{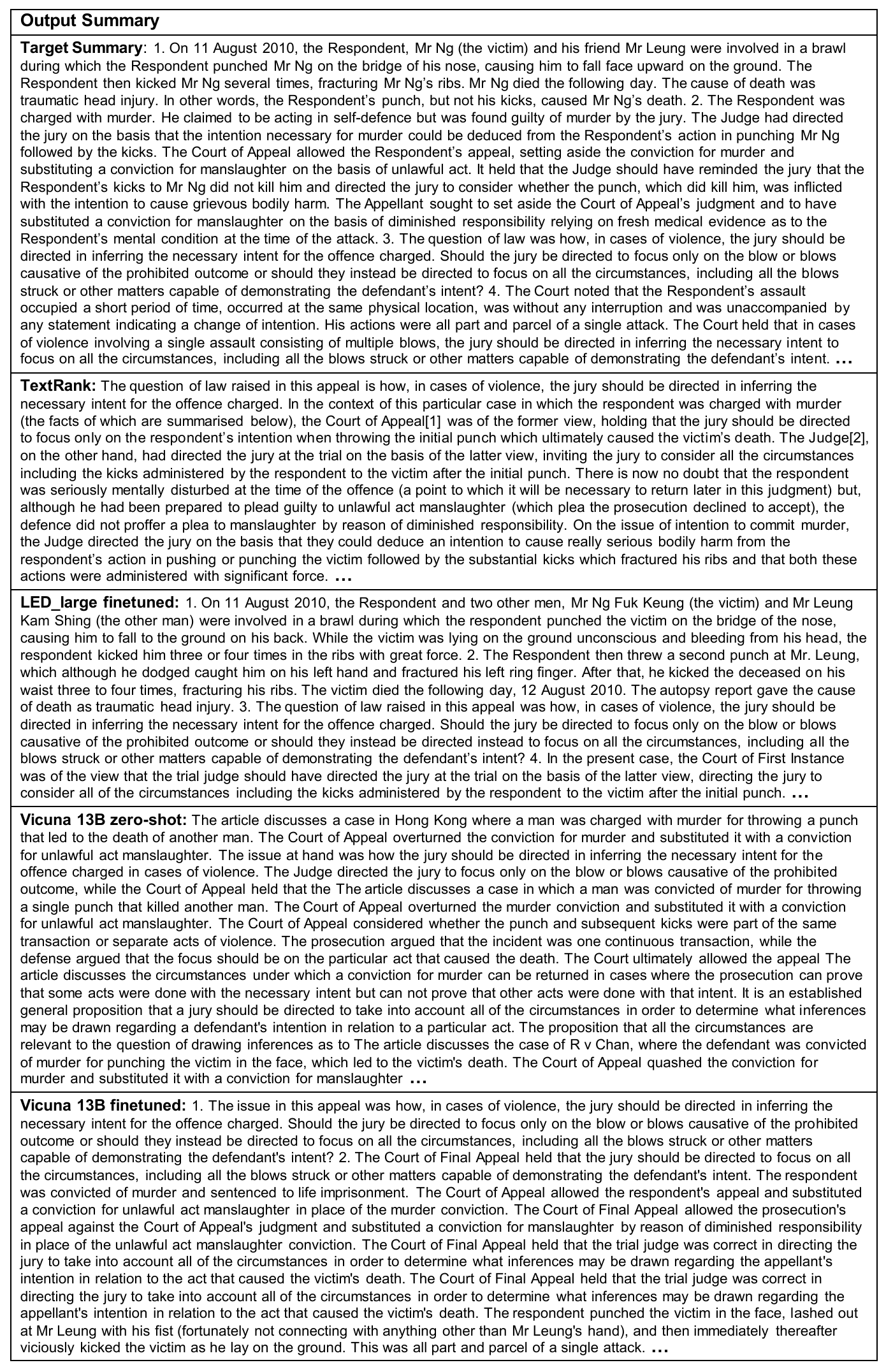} 
\caption{The output summaries of an example from the CLSum-HK.}\label{fig:clsum_hk_case_study} 
\end{figure*} 

\subsection{Case Study}
We conduct a case study to compare and analyze summaries generated by different models. 
Figure \ref{fig:clsum_hk_case_study} presents fragments in the target summary and different models' output summaries. 
When comparing these summaries, we find that summaries produced by extractive summarization methods lack fluency and readability. The summaries generated by LLM (Vicuna) under the zero-shot setting are quite different from the target summary in format. 
The summaries generated by the fine-tuned models (LED and Vicuna) are closer to the target summaries in both content and format, even though the models used here were fine-tuned with only 100 samples. 
This case study demonstrates that supervised fine-tuning is crucial for ensuring that the generated summaries meet the content and formatting requirements.

\section{Conclusion and Future Work}
\label{sec:conclusionandfuturework}
This paper introduces CLSum, a large-scale summarization dataset covering court judgments from four common law jurisdictions, including Canada, Australia, the United Kingdom, and Hong Kong SAR. 
Besides, we propose a large language model (LLM) based solution for the low-resource court judgment summarization. We present a series of methods to deal with three challenges: 1) training supervised summarization models with very limited labeled data, 2) identifying the salient content dispersed within the long judgment document, and 3) improving the efficiency of summarization models and training methods to process long input documents and summaries. 
This is the first court judgment summarization work adopting LLMs in data augmentation, summary generation, and evaluation. Specifically, we design an LLM-based data augmentation method named knowledge-constrained rephrasing. It introduces legal knowledge into the prompts of LLMs to constrain the synthesized text to accurately use legal concepts. 
Furthermore, we design a legal knowledge enhanced evaluation metric named LTScore to assess the generated legal text. 
We employ various summarization methods as baselines and benchmark them on our CLSum dataset. 
Our experimental results verify that the LLM-based summarization methods can perform well in the few-shot and zero-shot settings. Our LLM-based data augmentation method can alleviate the impact of low data resources.  
We carry out comprehensive comparative experiments to find essential model components and settings that are capable of enhancing summarization performance, including the training set size, foundation model architecture, SFT, and the RLHF process.

In future work, we aim to improve the dataset, summarization models, and evaluation metrics. We plan to build a larger dataset covering judgments from more jurisdictions. 
To further improve summarization models, we plan to use larger LLMs and design specific alignment mechanisms that adequately cater to the complex requirements in the legal field. 
We also plan to design more sophisticated evaluation methods to accurately evaluate the generated legal text. 
Overall, low-resource court judgment summarization remains an open problem, which offers the potential for improvement and requires novel solutions to address the associated issues. 

\bibliography{main}

\section*{Appendix}
\label{sec:appendix}

\subsection*{A1. Summarization Models' Few-shot Performance}
\label{sec:appendix_a1}

\begin{table*}[!htb]
\renewcommand\arraystretch{0.9}
\scriptsize
\centering
\setlength{\tabcolsep}{0.8mm}{
\caption{\label{autoeval:combined_summary_fewshot} Automatic evaluation results of summarization models’ few-shot performance. ``N examples" denotes using N examples when fine-tuning models.}
\begin{tabular}{lcccc}
\hline
\textbf{\multirow{3}*{Method}} & \multicolumn{4}{c}{\textbf{CLSum-CA}}\\
\cline{2-5}
~ & \textbf{10 examples} & \textbf{50 examples} & \textbf{100 examples} & \textbf{500 examples} \\
~ & \textbf{R1 / R2 / RL} & \textbf{R1 / R2 / RL} & \textbf{R1 / R2 / RL} & \textbf{R1 / R2 / RL} \\
\hline
$\mathrm{LLaMA_{7B}}$ & 46.41/15.95/19.00 & 45.61/15.80/20.08 & 47.91/18.10/20.74 & - \\
$\mathrm{LLaMA_{13B}}$ & 44.08/16.80/18.30 & 45.44/15.70/20.22 & 48.09/17.00/20.45 & - \\
$\mathrm{Vicuna_{7B}}$ & 47.94/16.78/20.77 & 47.02/17.00/21.64 & 47.62/17.36/22.05 & - \\
$\mathrm{Vicuna_{13B}}$ & 48.36/\textbf{17.50}/20.32 & 49.78/19.29/\textbf{22.72} & 50.66/19.22/\textbf{22.68} & - \\
LongT5 & 48.97/12.79/18.10 & 52.77/19.23/21.15 & 55.85/19.98/21.48 & - \\
$\mathrm{LED_{Base}}$ & 51.41/16.76/20.75 & 55.10/19.39/21.36 & 54.57/19.63/21.32 & - \\
Legal-LED & 51.94/16.65/20.80 & 54.63/19.10/21.43 & 56.04/20.33/21.73 & - \\
$\mathrm{LED_{Large}}$ & \textbf{54.37}/17.29/\textbf{20.85} & \textbf{56.27}/\textbf{19.52}/21.54 & \textbf{57.23}/\textbf{21.15}/22.65 & - \\
\hline
\textbf{\multirow{3}*{Method}} & \multicolumn{4}{c}{\textbf{CLSum-HK}}\\
\cline{2-5}
~ & \textbf{10 examples} & \textbf{50 examples} & \textbf{100 examples} & \textbf{500 examples} \\
~ & \textbf{R1 / R2 / RL} & \textbf{R1 / R2 / RL} & \textbf{R1 / R2 / RL} & \textbf{R1 / R2 / RL} \\
\hline
$\mathrm{LLaMA_{7B}}$ & 53.15/23.15/24.31 & 50.66/22.81/25.39 & 51.71/23.30/26.18 & - \\
$\mathrm{LLaMA_{13B}}$& 52.75/22.95/25.03 & 51.98/22.76/25.22 & 52.21/23.99/26.06 & - \\
$\mathrm{Vicuna_{7B}}$ & \textbf{55.58}/\textbf{25.57}/\textbf{26.26} & 54.84/25.26/26.50 & 55.01/25.26/26.42 & - \\
$\mathrm{Vicuna_{13B}}$ & 54.14/24.83/26.15 & \textbf{56.04}/\textbf{26.99}/\textbf{27.67} & 55.07/26.18/\textbf{26.78} & - \\
LongT5 & 51.35/19.38/21.39 & 55.36/24.81/23.50 & 56.29/\textbf{26.67}/24.85 & - \\
$\mathrm{LED_{Base}}$ & 53.03/21.36/21.89 & 53.62/23.52/22.65 & 55.56/25.47/23.04 & - \\
Legal-LED & 53.26/22.05/22.54 & 54.23/24.45/23.58 & 56.10/25.50/23.74 & - \\
$\mathrm{LED_{Large}}$ & 53.96/22.52/22.42 & 53.61/22.93/22.23 & \textbf{56.43}/26.49/24.92 & - \\
\hline
\textbf{\multirow{3}*{Method}} & \multicolumn{4}{c}{\textbf{CLSum-UK}}\\
\cline{2-5}
~ & \textbf{10 examples} & \textbf{50 examples} & \textbf{100 examples} & \textbf{500 examples} \\
~ & \textbf{R1 / R2 / RL} & \textbf{R1 / R2 / RL} & \textbf{R1 / R2 / RL} & \textbf{R1 / R2 / RL} \\
\hline
$\mathrm{LLaMA_{7B}}$ & 55.12/26.29/23.57 &  59.68/26.54/25.23 & 59.77/27.02/25.72 & 60.52/27.77/26.09 \\
$\mathrm{LLaMA_{13B}}$ & 60.70/27.38/25.47 & 59.61/26.37/25.51 & 60.83/\textbf{28.09}/26.26 & 61.26/28.54/\textbf{26.70} \\
$\mathrm{Vicuna_{7B}}$ & 58.80/26.87/25.40 & 59.77/27.26/25.92 & 60.05/26.98/25.50 & 61.74/28.58/26.68 \\
$\mathrm{Vicuna_{13B}}$ & 60.00/\textbf{28.07}/\textbf{26.02} & \textbf{60.91}/\textbf{28.18}/\textbf{26.11} & 60.42/28.01/\textbf{26.44} & 61.05/28.65/26.46 \\
LongT5 & 55.47/23.24/22.38 & 57.80/25.81/24.13 & 58.07/25.99/24.48 & 58.51/26.73/25.89 \\
$\mathrm{LED_{Base}}$ & 59.61/24.19/22.00 & 60.23/26.59/23.48 & 60.28/27.64/24.37 & \textbf{62.06}/\textbf{29.40}/25.30 \\
Legal-LED & \textbf{60.62}/25.56/23.45 & 60.71/26.55/23.78 & \textbf{61.52}/27.59/24.49 & 61.62/28.97/25.67 \\
$\mathrm{LED_{Large}}$ & 59.33/24.28/22.55 & 60.78/26.79/24.05 & 61.42/27.78/24.58 & 61.78/28.90/26.28 \\
\hline
\textbf{\multirow{3}*{Method}} & \multicolumn{4}{c}{\textbf{CLSum-AUS}}\\
\cline{2-5}
~ & \textbf{10 examples} & \textbf{50 examples} & \textbf{100 examples} & \textbf{500 examples} \\
~ & \textbf{R1 / R2 / RL} & \textbf{R1 / R2 / RL} & \textbf{R1 / R2 / RL} & \textbf{R1 / R2 / RL} \\
\hline
$\mathrm{LLaMA_{7B}}$ & \textbf{58.07}/27.92/28.29 & 58.11/29.28/30.22 & 57.08/29.32/30.53 & 57.77/30.69/30.81 \\
$\mathrm{LLaMA_{13B}}$ & 57.73/27.61/27.85 & 57.37/\textbf{29.49}/\textbf{30.67} & 59.19/\textbf{30.96}/\textbf{31.05} & 59.32/31.60/\textbf{32.16} \\
$\mathrm{Vicuna_{7B}}$ & 57.27/\textbf{29.65}/\textbf{29.20} & 56.80/29.42/29.49 & 57.27/30.01/30.40 & 57.66/30.13/30.47 \\
$\mathrm{Vicuna_{13B}}$ & 55.03/27.45/27.27 & 55.70/28.58/29.52 & 56.41/29.12/29.95 & 57.73/30.41/30.95 \\
LongT5 & 57.46/25.62/26.81 & \textbf{60.02}/29.24/28.14 & 60.58/30.31/29.18 & 61.47/31.21/30.40 \\
$\mathrm{LED_{Base}}$ & 57.18/24.73/26.04 & 59.03/27.57/27.46 & 59.68/29.22/28.24 &  61.91/31.60/30.21 \\
Legal-LED & 57.97/25.72/26.63 & 59.84/28.85/28.16 & 60.00/29.37/28.52 & 61.86/31.74/30.24 \\
$\mathrm{LED_{Large}}$ & 56.73/24.25/26.41 &  59.29/27.81/28.42 & \textbf{61.19}/29.90/29.22 & \textbf{62.77}/\textbf{32.07}/31.05 \\
\hline

\end{tabular}}
\end{table*}

Table \ref{autoeval:combined_summary_fewshot} presents various summarization models' few-shot performance. We fine-tune summarization models on training sets of increasing size (from ten examples to hundreds of examples). 
Figures \ref{fig:clsum_ROUGE2_f1_impact_of_training_sample_num}, \ref{fig:clsum_BARTScore_f1_impact_of_training_sample_num}, and \ref{fig:clsum_LTScoreAverage_f1_impact_of_training_sample_num} visualize the impact of training set size on the ROUGE-2 scores, BARTScore, and LTScore of the generated summaries. 
For relevant result analysis, please refer to Section \ref{subsec:summarization_results}.

\subsection*{A2. Summarization Models' Details}
\label{sec:appendix_a2}
These summarization models' implementation details are shown in Table \ref{table:model_details}.

\begin{table*}[t]
\scriptsize
\renewcommand\arraystretch{1.1}
\centering
\setlength{\tabcolsep}{0.8mm}{
\caption{Details of summarization models.}\label{table:model_details}
\begin{tabular}{lccccccc}
\toprule 
\textbf{Model}& \textbf{Architecture} &\textbf{Params} & \textbf{\makecell{Enc/Dec\\Layers}}& \textbf{Heads} & $\mathbf{d_{model}}$& $\mathbf{\mathrm{d}_{ff}}$ & \textbf{Input Len} \\
\midrule
$\mathrm{LED_{base}}$ & Enc-Dec & 161.8M & 6 & 12 & 768 & 3,072 & 16,384 \\
$\mathrm{LED_{large}}$ & Enc-Dec & 459.8M & 12 & 16 & 1,024 & 4,096 & 16,384 \\
Legal-LED & Enc-Dec & 161.8M & 6 & 12 & 768 & 3,072 & 16,384 \\
$\mathrm{LongT5_{base}}$ & Enc-Dec & 247.6M & 12 & 12 & 768 & 2,048 & 16,384\\
$\mathrm{BLOOM_{560M}}$ & Dec Only & 559M & 24 & 16 & 1,024 & 4,096 & 2,048\\
$\mathrm{BLOOM_{7B1}}$ & Dec Only & 7.1B & 30 & 32 & 4,096 & 16,384 & 2,048\\
$\mathrm{LLaMA_{7B}}$ & Dec Only & 6.7B & 32 & 32 & 4,096 & 11,008 & 2,048\\
$\mathrm{LLaMA_{13B}}$ & Dec Only & 13.0B & 40 & 40 & 5,120 & 13,824 & 2,048\\
$\mathrm{Vicuna_{7B}}$ & Dec Only & 6.7B & 32 & 32 & 4,096 & 11,008 & 2,048\\
$\mathrm{Vicuna_{13B}}$ & Dec Only & 13.0B & 40 & 40 & 5,120 & 13,824 & 2,048\\
\bottomrule
\end{tabular}}
\end{table*}

\end{document}